# Multimodal AI Systems for Enhanced Laying Hen Welfare Assessment and Productivity Optimization

**Daniel Essien[1], Suresh Neethirajan[1,2,*]**

[1]Faculty of Computer Science, Dalhousie University, 6050 University Avenue, Halifax, NS B3H 4R2, Canada

[2]Faculty of Agriculture, Dalhousie University, Truro, NS B3H 4R2, Canada

[*]Author to whom correspondence should be addressed. Email: sneethir@gmail.com

## Abstract

The future of poultry production hinges on a revolutionary paradigm shift: transforming subjective, labor-intensive welfare assessments into data-driven, intelligent monitoring ecosystems. Traditional welfare evaluation methods—constrained by human limitations and unimodal sensor dependencies—fail to capture the intricate, multidimensional nature of laying hen welfare in modern commercial environments. Multimodal Artificial Intelligence (AI) emerges as the critical breakthrough technology, orchestrating sophisticated fusion of visual, acoustic, environmental, and physiological data streams to unlock unprecedented insights into avian welfare dynamics. This comprehensive review synthesizes 130 peer-reviewed studies, revealing multimodal AI's transformative potential in laying hen welfare monitoring. Through systematic analysis of fusion architectures, we demonstrate that intermediate (feature-level) fusion strategies achieve optimal robustness-performance equilibrium under real-world poultry conditions, delivering superior scalability compared to early or late fusion approaches. Our investigation exposes critical implementation barriers: sensor fragility in harsh environments, prohibitive deployment costs, inconsistent behavioral taxonomies, and limited cross-farm generalizability that collectively impede widespread adoption. To overcome these challenges, we introduce two pioneering evaluation frameworks: the Domain Transfer Score (DTS) quantifying model generalizability across diverse farm conditions, and the Data Reliability Index (DRI) assessing sensor data quality under operational constraints. Additionally, we propose a modular, context-aware deployment framework specifically engineered for laying hen environments, enabling scalable integration of multimodal sensing technologies. This review establishes the scientific foundation for transitioning from reactive, unimodal monitoring to proactive, multimodal welfare systems, ultimately catalyzing the evolution toward precision-driven, ethically-conscious poultry production that harmonizes productivity with animal welfare imperatives.

## 1. Introduction

Poultry Welfare is increasingly recognized not just as an ethical concern but also a major determinant of productivity, product quality and sustainability in modern poultry systems. The "Five Freedoms" framework which emphasizes the birds' right to live free from hunger, discomfort, disease, fear and to express its natural behavior, has long served as foundational reference for animal welfare assessment [1]. However, its binary framing (freedom vs suffering) has faced criticism for lacking depth in understanding the complexities of welfare experiences.

As a progression, the "Welfare Quality" framework introduced by Keeling emphasized on four measurable principles: good feeding, good housing, good health and appropriate behaviour [2]. More recently, advanced welfare models like Five Domains framework have gained popularity. The 2020 update of the model [3] extends traditional physical and functional domains (e.g. nutrition, physical environment, health and behavior) by including the animals mental state as a fifth domain, thus providing a more holistic view of welfare. Likewise, the concept of Animal Agency has emerged, stressing the importance of providing animals with choice and environmental control, transitioning from merely avoiding suffering to actively promoting positive welfare experiences.

Beyond ethics, improved welfare correlates with reduced mortality, enhanced productivity, reduced disease outbreaks and better product quality. This is particularly evident in laying hens, where positive welfare has been linked to quantifiable improvements. For example, laying hens raised under the supervision of digital livestock systems produced eggs with improved fatty acid profile, including higher oleic acid and reduced n-6/n-3 ratios, which are considered healthier for human consumption [4]. Similarly, Sledgers et al. [5] noted that when poultry were raised at lower stocking density, they experienced reduced mortality and required fewer antibiotic treatments.

The importance of laying hen welfare is gaining momentum globally. Large corporations and governments are increasingly adopting cage-free systems. Several multinational companies have pledged to source only cage free eggs, while those that resist face scrutiny and consumer backlash. In Parallel, the European Union endorsed the phasing out of all cages for farm animals under the "End the Cage Age" Initiative, with some EU countries mandating full cage-free systems.

Although regions like Africa, Latin America and Southeast Asia still rely heavily on caged systems, due to economic and infrastructural constraints, international NGOs are working to improve these welfare standards in these regions. Stokes et al. [6] report that UK farmers are actively promoting positive welfare opportunities by providing resources on their farms designed to facilitate "good life opportunities", such as those related to comfort and pleasure. Using a resource tier framework, their study revealed that 63% of assessments achieved a score of "Welfare +" or higher. Simultaneously, consumer demand for ethically raised products is increasing [7]. Although developing regions are still catching up, recent trends showcase a positive global shift towards improved poultry welfare.

Managing poultry welfare, particularly in large scale operations, remain challenging. Issues include large flock sizes, environmental variability, difficulty in identifying subtle behavioral changes. Subtle welfare changes, such as changes in activity patterns, feather pecking or social withdrawal usually go unnoticed in large farms when visual monitoring is limited. Sakamoto et al. [8] highlighted the need to evaluate complex factors like contact dermatitis, thermal stress etc., emphasizing the importance of automated, precise monitoring tools. They also cite the need for optimal environmental standards like little quality and thermal comforts.

## 1.1 Poultry Stress Indicators

Stress in laying hens arises from both physiological and behavioural responses to stimuli or perceived threat that disrupts their homeostasis. Although acute stress responses may serve adaptive purposes, prolonged or repetitive exposure is associated with compromised welfare, reduced productivity, and increased disease susceptibility.

Stressors are broadly grouped into three types: physical (temperature extremes, high stocking densities, noise, and poor air quality), psychological (fear of humans, frequent handling, social instability) and nutritional (feed restrictions, water contamination and micronutrient imbalances). The cumulative effect of recurring stress can lead to feature pecking, aggression, mortality, reduced egg production and poor shell quality, all of which negatively impact welfare and farm profitability.

It is also essential to differentiate between acute and chronic stress responses due to their differing physiological signatures and welfare implications. Acute stress typically occurs due to short term disturbances such as sudden handling, loud noise or predatory threats. Physiological signs may include elevated heart rate, alpha-amylase activity and vocalization spikes. In contrast, chronic stress stems from extended exposure to poor housing, unstable environmental conditions or persistent overcrowding, while is manifest as elevated corticosterone levels, sustained heterophil-to-lymphocyte (H/L) ratio, reduced preening behavior, and long-term productivity declines.

Assessment of poultry stress is traditionally based on indicators from these four primary domains: physical, behavioral, production-based and physiological. Table 1 summarizes key indicators in laying hens across 4 main domains: Physical, Behavioural, Production Based, Physiological. However, the complexity of these indicators requires a multidimensional approach rather than unimodal analysis. For instance, behavioral measures, such as exploration, dustbathing, and preening, are extremely reliable indicators of positive affective states, showcasing positive affective state and environment adequacy [9,10].

In contrast, physical indicators such as plumage condition, leg health (e.g. lameness), body condition, and mortality offer tangible, quantifiable methods of welfare evaluation. Figure 1illustrates common physical manifestation of stress in laying hens, such as drooping wings, cloacal staining, and reduced movement, which may indicate underlying issues like digestive distress, health issues or thermal discomfort.

Table 1. Key Laying Hen Stress Indicators.

| Indicator | Visual | Acoustic | Thermal | Physiological |
|---|---|---|---|---|
| Body Weight [11–16] | ✗ | ✗ | ✗ | ✓ |
| Tibia Size [11–16] | ✓ | ✗ | ✗ | ✓ |
| Body Wounds [11–16] | ✓ | ✗ | ✗ | ✗ |
| Feather Condition [11–16] | ✓ | ✗ | ✗ | ✗ |
| Gait Score/Lameness [11–16] | ✓ | ✗ | ✗ | ✓ |
| Cleanliness [11–16] | ✓ | ✗ | ✗ | ✗ |
| Footpad Dermatitis [11–16] | ✓ | ✗ | ✗ | ✗ |
| Skeletal Integrity [11–16] | ✓ | ✗ | ✗ | ✓ |
| Reduced Activity [6,17–20] | ✓ | ✗ | ✓ | ✓ |
| Feather Pecking [6,17–20] | ✓ | ✓ | ✗ | ✗ |

| Vocalization Pattern [6,17–20] | ✗ | ✓ | ✗ | ✗ |
|---|---|---|---|---|
| Change in feeding [6,17–20] | ✓ | ✗ | ✗ | ✓ |
| Fearfulness [6,17–20] | ✓ | ✓ | ✗ | ✓ |
| Egg Production rate [21–23] | ✗ | ✗ | ✗ | ✓ |
| Egg Shell Quality [21–23] | ✗ | ✗ | ✗ | ✓ |
| Meat Quality [21–23] | ✗ | ✗ | ✗ | ✓ |
| Mortality rate [21–23] | ✗ | ✗ | ✗ | ✓ |
| Body temperature [15,18] | ✗ | ✗ | ✓ | ✓ |
| Heart Rate [15,18] | ✗ | ✗ | ✗ | ✓ |
| Heterophil-to-Lymphocyte Ratio (H/L) [15,18] | ✗ | ✗ | ✗ | ✓ |

Recent studies in poultry welfare have expanded the range of physiological biomarkers. In addition to Heterophil-to-Lymphocyte (H/L) ratio, corticosterone levels (measured in feces or feather), alpha-amylase activity, and heart rate variability (HRV) are being explored as non-invasive indicators of stress [24,25]. Feather corticosterone offers a noninvasive, chronic stress measure that reflects long term avian stress physiology [26]. In contrast, alpha-amylase are sensitive biomarkers for acute stress events, as it responds quickly to nervous system activation. HRV, though extensively studied in mammals, is increasingly used in poultry to detect autonomic nervous system imbalances, especially under thermal or handling stress.

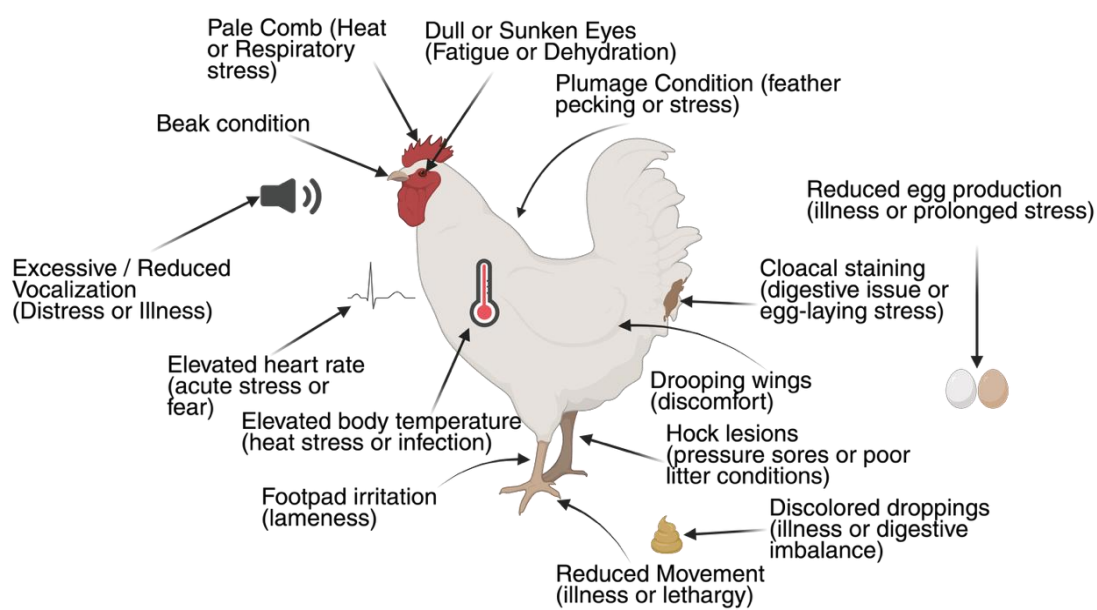

Figure 1. Key Physical Indicators of Welfare and Stress in Laying Hens.

Despite their widespread use, traditional welfare assessments methods such as manual inspection, observational scoring and production metrics are filled with limitations. These include subjectivity, infrequent application and labor intensiveness, which result in delayed or missed interventions, especially in large poultry farms [27]. These limitations emphasize the crucial need for more automated, objective, and continuous welfare monitoring systems.

## 1.2 Limitations of Single Sensor PLF and the Case of Multimodal AI

Precision livestock farming has emerged as promising framework for enhancing animal welfare, through continuous, real-time monitoring of health and behavior conditions of the animal [28]. Through the integration of various sensor technologies such as thermal cameras, microphones, video systems and environmental sensors, PLF can collect, process and interpret high resolution data streams to produce actionable insights and predictions [29,30]. In poultry systems, particularly laying hens, this framework solves lots of challenges especially in large scale operations.

However, while single-sensor systems have been widely adopted, they often fail to capture the full "contextual depth" needed for robust welfare assessment. In this context, contextual depth refers to the system's ability to synthesize information from various modalities (e.g. poultry vocalizations, gait or posture, behavior, and environmental data) to create a nuanced understanding of the emotional state and overall wellbeing of hen, similar to how humans integrate multisensory cues. These systems recognize the interplay between the different modalities, which single sensor systems lack. For example, an acoustic based system can detect distress but cannot differentiate whether it is caused by heat stress, disease or aggressive interactions [31].

While acoustic sensors have limitations in identifying the cause of stress, thermal imaging offers additional physiological insights by detecting heat-based anomalies such as fever or early signs of infectious diseases like Newcastle disease [32]. Yet thermal data alone may be compromised under different environmental temperatures, undermining generalizability across different farms. To address these constraints, Elmessery et al. [33] proposed a multimodal architecture that fused thermal images with RGB imagery to detect pathological phenomena such as Diseased eyes, Lethargic Chickens and Stressed Chickens. This multimodal approach not only compensates for the limitation of single modality, such as poor lighting or background thermal noise, but also broadens the dimensionality of diagnostic input, providing a more nuanced welfare assessment. Notably, the multimodal approach got a 97% F1 score, demonstrating the superiority of cross-modal data integrations for complex, real-world poultry environments.

Although Elmessery et al. [33] showcased impressive results, its applicability in real world poultry environments are still questionable. They are clearly discrepancies between high accuracy and the known operational challenges (e.g sensor drift and occlusion). Although multimodal showcase higher accuracy, to make a strong stance on multimodal strength in real world application, it needs to be validated across multiple housing systems, weather conditions and fluctuating environmental focus. If multiple validations aren't made, its strengths can't be certifiable.

To further illustrate, Table 2 compares unimodal and multimodal systems across several welfare monitoring applications. A key observation is the large amount of unimodal studies, particularly in poultry research. Despite its dominance, multimodal integration is consistently recognized as a crucial future direction and a solution to the inherent limitation of single modality systems.

Table 2. Comparative Performance of Unimodal and Multimodal Systems for Livestock Monitoring: Implications for Poultry Welfare.

| Application | Species | Modality | Sensors / Modality Used | AI Model/ Fusion Strategy | Performance Metrics | Key Observations | Reference |
|---|---|---|---|---|---|---|---|
| Chicken Counting | Poultry | Unimodal | IP ColorVu Lite Surveillance Cameras | YOLOv8 | Precision: 93.1%, Recall: 93.0% | High precision and Recall, even with occlusion and dense flock condition<br><br>Lays foundation for innovations in agriculture automation and welfare monitoring | [34] |
| Egg grading and Defect Detection | Poultry (Laying Hens) | Unimodal | Computer Vision | RTMDet (CNN-based for classification),<br><br>Random Forest (for weight prediction) | Accuracy: 94.8% (classification),<br><br>R2: 96.0% (weight prediction) | Serves as a solution for quality control of eggs<br><br>Utilizes deep learning for classification and regression for weighing | [35] |
| Cattle Welfare | Cattle | Multimodal | Surveillance cameras (Hikvision ), | YOLO + DeepSORT tracking with | Environmental: 95.0%, | Highlights an integrated, high- | [36] |

| | | | Environmental sensors (temp, humidity, CO2, light), Farm feeding database (DM, ADF, NDF, CP, Ca, P) | multimodal backpropagation Neural Networks and adaptive fuzzy logic (Mamdani inference, Gaussian MFs). | Feeding: 100%, Behavior: 93.6% (validation). All modules >90% accuracy; high agreement with manual evaluation. | accuracy welfare assessment framework using fuzzy logic for multimodal fusion. Methodology is highly useful for future poultry applications | |
|---|---|---|---|---|---|---|---|
| Anticipatory Behavior Detection / Flock Monitoring | Poultry (Chickens) | Multimodal | Accelerometers, Acoustic Recordings, Machine Vision | Unspecified | No specific metrics; Highlights high temporal sensitivity and anomaly detection. | Holistic welfare approach for poultry welfare monitoring. Although it lacks quantitative validation data, it provides a strong conceptual basis for poultry welfare assessment. | [37] |

As shown in Table 2, demonstrates that while unimodal models such as YOLOv8 and RTMDet achieve high precision and accuracy, their scope remain limited, focusing on isolated features such as object detection or classification. In contrast, multimodal systems combine environmental, feedings and behavioral data to deliver a more holistic and robust assessment of animal welfare. These systems not only deliver high performance metric (>90% accuracy) but also showcase high concordance with manual evaluations. Although poultry-specific multimodal studies are limited, the findings summarized in Table 2 provide a valuable conceptual framework for future.

Nonetheless, these high-performing multimodal systems come with tradeoffs. The fusion of heterogenous data streams, such as high frequency chicken vocalizations and low frequency environmental readings, often create temporal misalignment issues. Additionally, sensor calibration drift between sensors (e.g. RGB and thermal) may degrade fusion quality over time, requiring occasional recalibrations to maintain reliability. Equally, computational challenges such as real-time inference latency increase proportionally with the number of modalities and the complexity of the fusion model architecture. Additionally, interpretability also hinders real world adoption, as farmers and practitioners require clarity in order as to why the models made certain predictions. The inception of explainable AI is beginning to address this but most fusion models are yet to adopt this.

### 1.3 Review Scope and Objective

This review aims to critically examine the current findings and advancements made in Multimodal AI for poultry welfare, with a focus on laying hens. This review synthesizes recent advancements in multimodal AI and data fusion strategies to evaluate its effectiveness in enhancing welfare assessment, productivity, and operational efficiency. We explore the architectures, advantages, limitations and applications of Multimodal AI Systems for Poultry Farming. In addition, we address key technical and ethical challenges, including data quality issues, algorithms bias, and regulatory considerations.

### 1.4 Systematic Review Methodology

### 1.4.1 Review Framework and Guidelines

This systematic review is structured to establish a clear flow of information, from understanding the key concepts to presenting the latest findings and technologies and finally highlighting the challenges and future directions. The systematic review was conducted using the Preferred Reporting Items for Systematic Reviews and Meta-Analyses (PRISMA 2020) [38], guidelines to ensure transparency, methodological transparency, reproducibility and scientific merit. Although this review followed the PRISMA 2020 guideline for systematic review, the protocol was not registered in a public database like PROSPERO. Its absence is recognized as a limitation regarding transparency. The detailed selection process is visually summarized in Figure 2 in accordance with PRISMA 2020 guidelines.

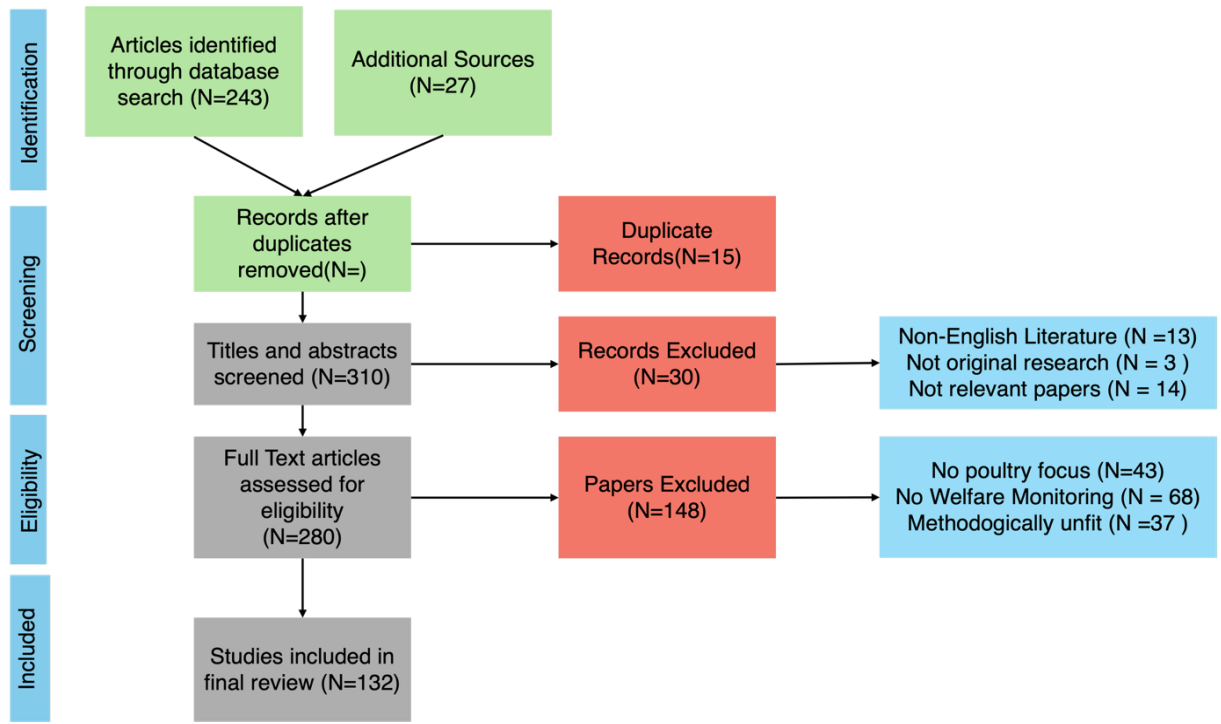

Figure 2. PRISMA (Preferred Reporting Items for Systematic Reviews and Meta-Analyses) 2020 flow diagram showcasing the methodological approach used in this systematic review.

### 1.4.2 Literature Search Strategy

The literature search was conducted using four electronic databases (Web of Science, Scopus, Google Scholar, and ScienceDirect) to ensure comprehensive coverage of peer reviewed articles across various domains. The keyword strategy was both inclusive and domain specific targeting four major themes: multimodal data fusion techniques, poultry-specific applications, sensor technologies, and environmental/behavioral monitoring.

This review focused on studies published from 1st Jan 2019 to 15th May 2025 to reflect the most recent developments in multimodal laying hen monitoring, including deep learning architectures (e.g. transformers), multimodal sensor fusion, wearable and noninvasive sensing systems, and intelligent automation in poultry farming. Selective inclusion of pre 2019 were used only if they presented foundational methodologies in machine learning, sensor integration or early poultry welfare frameworks.

Search strategy involved combining and permuting keywords: "laying hens", "precision livestock farming", "machine learning in poultry", "vocalization analysis", "acoustic and visual monitoring", "multimodal data fusion", "multisource" and "poultry welfare assessment".

The search strings used in each database followed this Boolean structure: ("laying hens" OR "egg-laying poultry") AND ("precision livestock farming" OR "smart farming") AND ("machine learning" OR "deep learning") AND ("sensor" OR "acoustic" OR "vision" OR "multimodal") AND ("welfare" OR "productivity"). The Boolean logic ensured a balanced inclusion of technical and domain specific studies. The exact search strings and search dates for each database is presenting in Table 3

Table 3. Database, Search Strings and Data of Search

| Database | Search String | Date of Search |
|----------|---------------|----------------|
| Web of Science | "multimodal AND poultry", "sensor fusion AND laying hens", "machine learning AND welfare" | 18th April 2025 |
| Scopus | ("laying hens" OR "egg-laying poultry") AND ("precision livestock farming" OR "smart farming") AND ("machine learning" OR "deep learning") AND ("sensor" OR "acoustic" OR "vision" OR "multimodal") AND ("welfare" OR "productivity") | 21st March 2025 |
| ScienceDirect | "multisource data AND layer hens", "automated monitoring AND poultry welfare" | 19th May 2025 |
| Google Scholar | "laying hens AND machine learning", "multimodal sensors AND poultry", "deep learning AND behavior" | 8th May 2025 |

### 1.4.3 Inclusion and Exclusion Criteria

Eligible studies were those conducted in English, published in peer reviewed journals with a focus on monitoring poultry (laying hen) behavior, welfare, productivity and health. This selected studies either utilized multisource, multimodal or unimodal methods to classify or quantify laying hen metrics such as feeding, activity levels, behaviors, disease. Grey literatures were selectively included, preprints were allowed due to the provision of substantial methodological detail and alignment with inclusion criteria. However, conference papers, thesis, technical reports and non-peer reviewed white papers were omitted to maintain a consistent level of methodological rigor and peer-reviewed standard.

Studies that were excluded were studies not in English, had no relation to poultry or laying hens, lacked the application of machine learning, or failed to include poultry science or welfare. Additional exclusions were made for papers that lacked methodological clarity or could not be reproduced due to insufficient reporting.

### 1.4.4 Screening and Reviewer Agreement

To ensure consistency and reduce bias, the retrieved articles were downloaded and reviewed by the lead author. The shortlist of studies was then sent to a subject matter expert (the supervising author) for secondary screening and validation. This two-step process help ensure relevance, credibility and alignment to research objectives. Any disagreements were solved through discussions

To ensure transparency and reliability, two independent reviewers screened a subset of 30 full-text articles. The inter-reviewer agreement between the lead author and a second independent reviewer was high with a Cohen's Kappa Cooeficient of 0.82, indicating significant agreement between both reviewers. Any differences were addressed through dialogue to reach final inclusion decisions.

The screening followed three steps, Firstly, a total of 243 papers were collected from through databases (N = 243) and manual search (N = 27). After removing 15 duplicate records, 310 papers remained for title and abstract screening. 30 papers were excluded due to lack of relevance, originally or not written in English. 280 studies paper proceeded to full text review. During this stage, 148 additional papers were excluded: 68 were on topics related to AI and Poultry monitoring, 43 did not focus on animal science and 37 were deemed methodologically unfit.

### 1.4.5 Data Extraction

Data extraction was conduction by the lead author using a standardized structured spreadsheet developed in Microsoft Excel. The extraction was tested using a subset of five included studies and iteratively refined to ensure completeness and relevance during further data extraction. Key data fields that were extracted include publication year, study objectives, animal species, sensor modalities used, AI model architecture, evaluation metric, outcomes and limitations.

Due to the heterogeneity of the selected studies, no meta-analysis was conducted. However, a descriptive synthesis was conducted to highlight methodological patterns, applications, use cases, and technological trends. The outcome of this critical synthesis recognized limitations and discussions for further research.

### 1.4.6 Risk of Bias

Risk of bias was assessed using the Newcastle-Ottawa Scale (NOS), a validated tool suitable for evaluating non-randomized studies in systematic reviews. Each study was independently evaluated across three domains: Selection (maximum of 4 stars), Comparability (maximum of 2 stars) and Outcome (maximum of 3 stars). For each study, a maximum of 9 stars could be assigned, with higher scores representing lower risk of bias.

The NOS evaluation was performed across all included studies, using explicit criteria and domain-specific indicators to ensure consistency. Studies scoring 7-9 were grouped as low risk bias papers, those scoring 4-6 stars as moderate risk and below 4 stars as high risk. Out of the 130 included studies, 43% were rated as low risk, 45% were rated as moderate risk and 12% as high risk.

### 1.5 Research Questions

This review is guided by the following research questions, developed to evaluate the scope, applicability and limitations of ethical implications of multimodal AI systems in enhancing laying hen welfare and productivity in commercial farms.

- How can multimodal AI systems improve the detection, monitoring, and management of laying hen welfare compared to unimodal systems
- What are the strengths and limitations of different sensing modalities when applied in large scale, real world laying hen farms
- Which multimodal fusion strategies are most suitable for real time welfare monitoring in poultry systems, while considering the limitations
- What are the practical and infrastructural barriers that constrain the deployment of multimodal AI technologies in poultry farms, and what are the potential solutions
- What are the ethical and welfare considerations associated with continuous welfare monitoring, particularly in relation to stress, animal anatomy and technology.

## 2. Multimodal AI Architecture and Fusion Strategies

## 2.1 Introduction to Fusion for Precision Poultry Farming

Precision poultry farming relies on the utilization of different data modalities such as visual, acoustic, thermal, and environmental data to monitor hen welfare, health and behavior. Despite the technological maturity of individual sensors, current implementation often treats each modality in isolation. This fragmented approach limits contextual understanding, increases vulnerability to sensor-specific noise, and reduces generalizability.

Unimodal systems, while effective under controlled facilities, often degrade in commercial environment. For instance, video-based systems are highly sensitive to occlusion caused by flock density, equipment and particulate interference, which restricts its reliability in real world farms [39]. These vulnerabilities underscore the crucial need for multimodal approaches, which combine complementary data sources to enhance system resilience and fault tolerance even when one modality fails.

Derakhshani et al. [40] utilized both video and inertial measurement unit (IMU) data but applied the video stream only for behavioral labeling, while relying on IMU data for classification. This modality segregation missed the opportunity to perform data-level or feature-level fusion, in turn failing to leverage cross-modal correlations that could enhance model robustness and sematic fidelity. In contrast, Kate et al. [41] , showcased the power of multimodal fusion by integrating acoustic, video and biometric sensor data to achieve a more robust, contextually aware understanding of animal's welfare, a principle directly transferable to poultry welfare.

Multimodal Fusion has showcased significant success across domains such as human activity recognition, emotion recognition, medical application [42–44]. These applications often employ deep fusion architectures like convolutional neural network (CNN) with shared intermediate layers, attention-based transformers, and graph-based models, to learn joint feature representation cross modalities.

Emerging transformer-based architectures such as Perceiver IO and Flamingo have redefined the capabilities of cross modal learning by offering superior scalability and capability to process unstructured, asynchronous input streams. These models dynamically handle relevant modality-specific inputs without strict temporal alignment, an advantage that works well in real, noisy data gotten from poultry farm sensors. In such systems, acoustic sensors may collect data at kHz rates while environmental sensors may collect data update hourly. A fused architecture with cross-modal attention and time-aware gating permits the integration of multi-temporal data, enabling the detection of progressive welfare issues like overheating or behavioral suppressions.

Another major issue that is often overlooked is the temporal resolution and sampling mismatch between various modalities. Acoustic sensor typically possess millisecond-level resolution and continuous data collection, enabling granular data collection of vocalization and stress calls, whereas infrared thermography (IRT) often operates in lower frame rates (~1fps) in short bursts, while environmental sensors that capture $CO_2$ and ammonia operate on the scale of minutes or hours due to slow diffusion dynamics of gases and sensors refresh rate. The difference in sampling windows and temporal resolution differences are illustrated in Figure 3. This temporal inconsistency creates alignment issues for the multimodal fusion systems, where synchronous inputs are required for feature concatenation. If these temporal differences are not properly addressed, they can lead to inaccurate predictions or complete model failure.

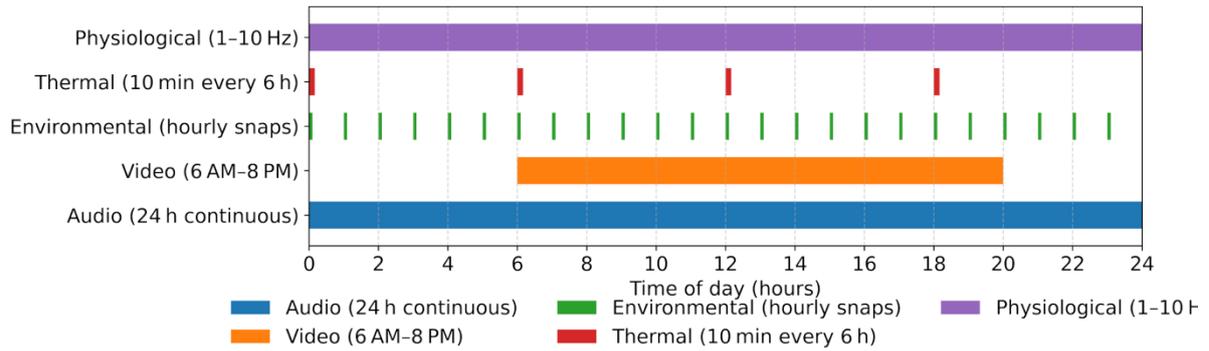

Figure 3. Sampling schedules and temporal resolution of multimodal sensors in precision agriculture over a 24-hour cycle.

Data fusion has showcased immense improvement in system robustness, particularly under sensor degradation, missing data, or unclear input signals. Although its implementation in poultry systems remain scarce it offers a more robust poultry system. For example, a multimodal system that combines audio-based stress signals with thermal fluctuations to pre-emptively identify overheating or aggression events might outperform a unimodal alternative.

In other areas of PLF, multimodal architectures have shown potential in disease detection (e.g. cow lameness [45], behavioral tracking [46] and welfare monitoring [36]). However, widespread application of these techniques remains limited, primarily due to the inherent complexities of high animal density leading to occlusions, data annotation challenges, rapid growth cycles of poultry and high cost of system implementation relative to poultry's narrow profit margins.

The observed absence of extensive studies is surprising, given the well documented limitations of unimodal systems in poultry farms such as inherent lack of context, poor fault tolerance, and varying environmental conditions. A sophisticated multimodal fusion framework could theoretically alleviate these issues, yet not much research has been done within the poultry domain. At the time of this study, no publicly available poultry specific multimodal dataset exists that integrates even two sensing modalities, thereby emphasizing both the gap and opportunity.

## 2.2 Categories of Fusion: Early, Intermediate, and Late

Multimodal fusion is a transformative technique in precision poultry farming (PPF), enabling the integration of heterogenous sensor data such as acoustic, visual, thermal, and environmental, to detect diseases, assess welfare and optimize poultry farm operations. Fusion strategies are typically categorized by the stage which the data streams are fused: early (low-level), intermediate (feature-level), and late (decision-level) fusion (Figure 4). Each approach has unique tradeoffs in information richness, robustness to noise, computational efficiency and real-world feasibility.

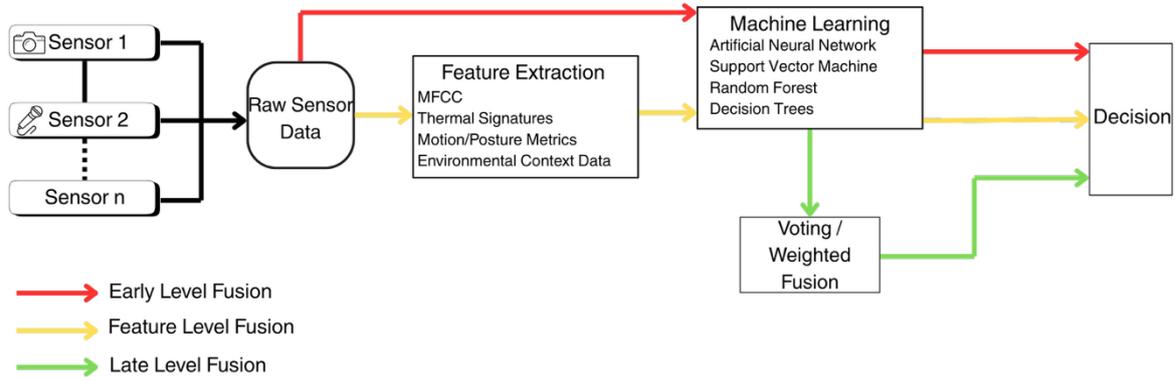

Figure 4. Illustrative Overview of Fusion Strategies: Early, Intermediate and Late Fusion

### 2.2.1 Early Fusion (Low-Level)

Early fusion involves the direct concatenation of raw data from each modality before feature extraction. Although this approach retains the highest degree of modality-specific detail (e.g. temporal and spatial signals), it is often impractical in poultry systems. The diverse nature of farm sensors (e.g. different sampling rates, varying spatial resolution, asynchronous data capture) introduce severe alignment challenges that can deteriorate model performance and make deployment tasking. Early fusion requires all raw sensor data to be transformed into a single multidimensional tensor, which is ingested by multimodal architectures capable of handling high dimensional fused inputs, such as 3D CNNs, ConvLSTM or transformer-based encoders. This process requires data normalization across modalities and careful management of channel depth (e.g. stacking RGB, thermal and spectrogram).

Despite its benefits, early fusion requires sensors to capture data at the same time and place, which is difficult is poultry farms due to factors like lighting variations, dust interferences and inherent poling rates of diverse sensors. Martin et al. [47] noted that even minor temporal misalignments or presence of modality specific noise (e.g. static sounds from microphones or glares in video feeds) can lead to corrupted joint representations.

To address this, early fusion systems requires extensive data alignment and interpolation pre-processing. While feasible in lab settings, it possesses high computational overhead and introduces errors when deployed in resource constrained edge devices [48]. Engineers must employ robust techniques such modality masking or adaptive resampling layers to simulate and minimize sensor drift or loss during training [49].

Recent studies in multimodal learning are modernizing early fusion by using learnable alignment modules, temporal transformers, or signal warping layers to reduce reliance on strict synchronization [50]. These could theoretically allow raw multimodal data to be fused earlier on, while preserving fine grain signals. However, such innovative solutions remain largely invalidated and benchmarked in poultry research.

Pawłowski et al. [51] demonstrated this limitation in a non-agricultural domain, where late fusion (accuracy = 0.969) surpassed early fusion (0.940) under noisy real-word scenario. These results directly challenges the assumed dominance of early fusion's granularity, highlighting its sensitivity to real-world applications. Additionally, controlled studies that report high F1 scores and accuracy using early fusion often fail to mention deployment challenges, raising questions about external validity and transparency

Several studies report high accuracy and F1 scores of early fusion models yet simultaneously acknowledge issues like noise, latency and temporal resolution mismatch. It becomes unclear whether the high accuracy comes from genuine sensor synergy or from overfitting of tightly controlled datasets. Without standardized evaluation protocols, such claims must be interpreted cautiously.

While currently underutilized in poultry systems, future works should focus on enhancing early fusion by utilizing cross-modal transformers or signal warping layers, capable of resolving temporal misalignment. These innovations would boost early fusion's potential for capturing those granular, instantaneous welfare cues. However, until early fusion systems are properly validated in real farm environmental, claims about early fusion efficiency in poultry farming should be viewed cautiously with critical attention given to its operational fragilities.

### 2.2.2 Intermediate Fusion (Feature-Level)

Intermediate (Feature-Level) Fusion extracts modality specific features before combining them into a unified latent representation. This method is well suited for poultry systems where data heterogeneity and sensor dropouts are common. Intermediate fusion is often executed using modality-specific neural encoders (e.g. CNNs for images, RNNs for audio, or MLP for structured environmental data) which transforms raw input into a vector space and then fused, via concatenation or learned attention, into a joint embedding that represents the full context of the hen's full condition.

Compared to early fusion, intermediate fusion is known for its flexibility. Modalities can be processed at different rates, which is valuable in farm conditions where sensors may fail or lag. Attention mechanisms such as cross-modal transformers [52] help learn soft alignment across modalities, particularly when signals are weakly correlated or asynchronous.

Additionally, contrastive loss functions (e.g. InfoNCE or triplet loss) are used to train these systems effectively, to enforce semantic consistency between modalities and enhance the model's ability to differentiate between normal and abnormal hen patterns [53]. Canonical Correlation Analysis (CCA) and its neural extensions are also utilized to maximize alignment between embedding spaces [54], allowing the system to understand interpretable relationship (e.g. how increased ambient noise relates to stress indicators in videos)

However, this approach discards raw signal nuances that carry subtle welfare indicators and cross modality information, such as micro changes in thermal patterns or low amplitude distress calls. Thus, while feature-level fusion strikes a balance between robustness and efficiency, it can lose those cross-modality nuances in welfare monitoring. Despite this, feature-level fusion remains prevalent in literature across animal multimodal sensing literature [55], due to its generalizability and reduced sensitivity to misalignment.

Its popularity likely occurs more from an engineering feasibility standpoint rather than optimal information retention. For example, lightweight modality-specific encoders deployed at the edge, and their embeddings fused centrally reduces data transmission volume and improves real time performance in real farm conditions. Yet, this architecture may still miss cross-modal dynamics unless the fusion layer explicitly models interactions, such as synchronized changes in posture and vocalization during early signs of stress.

Although intermediate fusion appears to be a pragmatic solution, it may possess a false sense of robustness if evaluated solely on performance metrics without accounting for what information was lost or skewed during embedding. The feature extraction which occurs in each modality, prior to fusion, introduces an abstraction layer that may dilute the subtle intermodal interactions which are essential for accurate welfare assessment.

Future works should consider hybrid designs that combine feature level fusion with residual raw input branches or signal reconstruction losses. Additionally, Matrix-based attention layers and cross-modal transformers can preserve granular signals while maintaining scalability

### 2.2.3 Late Fusion (Decision-Level)

Late fusion processes each modality independently using dedicated models and combines their output via ensemble methods such as voting, averaging or meta modelling [56]. Its modular architecture makes it suitable for real-time, scalable applications, where plug-and-play sensor modules and fault-tolerant pipelines are ideal.

However, this independent approach is counterintuitive. Decision level isolation restricts the model from learning intermodal dependencies, such as correlations between abnormal posture (detected via vision) and elevated distress vocalizations (captured via audio data), both which, when simultaneously occurring might indicate early signs of disease or aggression. Furthermore, there is an absence of gradient-level interactions across modalities in decision fusion, meaning no joint optimization occurs over the modality space [57]. This makes it inappropriate for capturing cross-modal co-activations or synergistic patterns critical in welfare assessments.

Late fusion typically outputs independent scalar prediction (e.g. class probabilities), which are then aggregated using ensemble strategies like weighted averaging, stacking or rule-based voting. Although simple, these methods ignore the underlying structure of each modality's output (such as attention maps or temporal embeddings), thereby disregarding potentially useful contextual information. Such architectures often fail to capture high-order fusion patterns or perform matrix-level reasoning.

Though limited, a speculative example in poultry could involve a late-fusion model combining video-based behavior recognition and microphone detected distress calls to flag potential agnostic behaviors, with each modality analyzed by independent lightweight models. While this system lacks intermodal learning, its modular design is resilient and possess fault tolerance.

Currently, there are limited poultry focused studies that utilized or benchmarked fusion strategies against one another in real farm conditions, leaving a major gap in our understanding of which approaches suit the practical demands of poultry welfare assessments. To date, no study has directly compared early, intermediate, and late fusion techniques under poultry specific stressors like high dust density, low lighting, and ambient noise.

In addition, emerging paradigms in multimodal AI such as attention-based fusion, dynamic modality weights and transformer based cross modal architectures have yet to be explored in poultry systems, despite their proven effectiveness in other domains like healthcare, human activity recognition, and autonomous driving. Future research should conduct multimodal fusion studies in poultry systems and then explore hybrid fusion frameworks capable of dynamically adjusting fusion strategies based on context, sensor reliability, and environmental volatility.

The structural simplicity of late fusion model reduces its ability to understand cross modal cues. The lack of joint optimization and gradient level interaction somewhat contradicts the main purpose of utilizing a multimodal approach for poultry welfare. Additionally, the model used in each modality pipeline equally matters. Differences in architectural depth, input preprocessing and temporal resolution can affect the information that is extracted and passed on for fusion. If each model extracts features with inconsistent levels of details, the final prediction will reflect this inconsistency.

## 2.3 Advantages and Limitations of Fusion Strategies

Each multimodal fusion strategy presents unique strengths and weaknesses when applied to poultry system, especially in environment characterized by sensor noise, asynchronous data streams and various hen behaviors. Early fusion, while preserving data integrity, suffers from synchronization challenges and increase noise propagation due to direct concatenation of raw inputs. Feature Level fusion offers a balance between data integrity and computational efficiency but may still disregard some subtle cross modal dependencies critical for subtle welfare monitoring. Late fusion, although modular and scalable, doesn't have the ability to model intermodal interactions prevention the detection of complex, temporality align behavioral cues. Table 4 summarizes the tradeoffs between these fusion strategies.

Table 4. Comparative Summary of Multimodal Fusion Strategies for Laying Hen Monitoring

| Dimension | Early Fusion | Intermediate Fusion | Late Fusion |
|---|---|---|---|
| Modality Synchronization Requirements | **High**: Requires temporal and spatial alignments, which is difficult in poultry environment and may lead to semantic confusion or degraded performance | **Moderate**: Each modality is encoded before fusion, reducing synchronization constraints. However, minimal alignment is required but preprocessing can reduce inconsistencies | **Low:** Each modality is processed separately. Ideal for loosely coupled data sources |
| Gradient Flow Across Modalities | **Shared**: A unified network fuses the input, allowing backpropagation across all modalities. | **Partially Shared:** Gradients propagate through modality specific sub-networks and a shared fusion module. | Independent: Each modality is trained independently, and outputs are merged after. While gradients don't flow across modalities, this approach enhances explainability and fault tolerance. |
| Robustness to Sensor Noise or Failure | **Low:** Noise from one modality essentially corrupts the shared representation | **Moderate:** Partial feature extraction provides robustness. Feature degradation of one modality won't affect entire system | **High:** As decisions are made independently, failure in one modality has minimal impact in overall system functionality. |

| Ability to Capture Cross-Modal Relationships | **High:** Fusing data at granular level aids in cross modal attention and joint laten space learning but success depends heavily of temporal synchronization and noise control | **High:** Has tendency to capture intermodal relationship through feature embedding. | **Low-Moderate:** Limited inter modality learning due to isolated processing. |
|---|---|---|---|
| Scalability for Real-Time Farm Environments | **Moderate to Low:** Increased computational cost as input modalities increase. | **Moderate:** suitable but remains computationally expensive with many modalities | **High:** its modular nature makes it suitable for edge or low power devices |
| Typical Use Cases in Precision Livestock Farming | Disease detection Stress monitoring | Health diagnosis Individual tracking | Farm wide monitoring |
| References | [58–60] | [58,60,61] | [58,60,61] |

## 2.4 Proposed Framework for Multimodal & Sensor deployment in Real Farm Scenario

To address the practical limitations of sensors-based and multimodal AI systems in commercial poultry environments, we propose a context-aware, modular, scalable framework designed for adaptability, ease of integration and robustness (Figure 5). At its core, the framework relies on non-invasive and distributed sensors, including overhead cameras, directional microphones, thermal cameras, and environmental sensors. These are strategically positioned to minimize occlusion and biosecurity risks while maximizing space and sensor coverage.

To minimize data loss or signal degradation, the system uses multiple data types and overlapping coverage. This ensures that if one sensor fails or produces inconsistent data, others are still useful. Edge device with lightweight AI models performs on-site preprocessing, anomaly detection, event tagging and possibly feature extraction to reduce data volume and latency.

To enhance real-time responsiveness, edge-level fusion modules such as lightweight gated recurrent units (GRU) or temporal convolutional networks (TCNs) can be infused to capture short-term fluctuations in behavior or environments. GRU, a type of recurrent neural network, can capture sequential patterns and time-dependent behaviors with little computational load. For example, GRUs can capture 30-second clips of acoustic sounds and thermal images to detect pre-aggression, alerting end-user for early intervention before escalation, improving latency performance and reducing burden on centralized system.

Central to the framework is the cloud-based data fusion layer, where the various modalities are integrated depending on the chosen fusion strategy. Insights and predictions are now sent down to a graphical user interface (GUI) or Mobile Application provided for farmers with actionable intervention recommendations.

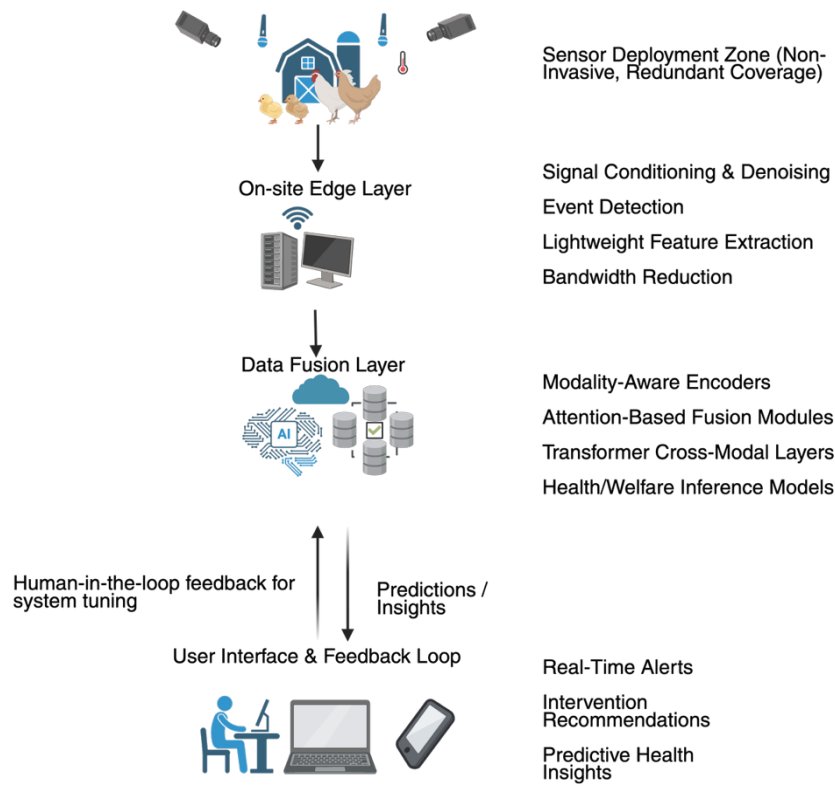

Figure 5. Modular Framework for Multimodal AI in Poultry Farms.

This layered, modular architecture offers scalability across different farm sizes and configurations, and it is adaptable to different environmental conditions and compatible to various sensor brands. These frameworks create replicable blueprint for deploying intelligent, welfare centric multimodal poultry systems.

## 3. Sensing Technologies

Precision monitoring of laying hen welfare requires sensor-based systems capable of capturing visual, acoustic, environmental, and physiological data. However, these technologies differ significantly in terms of accuracy, cost, scalability, and robustness. Although visual and acoustic sensors are widely used for behavior and disease detection, their effectiveness is often hindered by occlusion and background noise. Alternatively, physiological sensors offer individualized stress metrics but are invasive and hardware sensitive, whereas environmental sensors shine in environmental monitoring but can't identify behaviors.

This section provides a detailed comparison of the four sensing modalities, followed by an assessment of its deployment feasibility in commercial poultry setting. Table 5 highlights the core advantages and limitations of each modality.

Table 5. Comparative Summary of Sensing Modalities for Laying Hen Welfare Monitoring

| Modality | Strengths | Limitations | Best Use Cases |
|---|---|---|---|
| Visual | Rich spatial context, Supports behavior and posture analysis | Occlusion issues, Affected by lighting variations, High data volume | Gait, activity recognition, anomaly detection |

| Acoustic | Non-invasive,<br><br>Effective for early disease and distress detection. | Affected by ambient noise,<br><br>Overlapping vocalizations in chicken,<br><br>Limited Spatial context | Respiratory health, acoustic distress signals |
|---|---|---|---|
| Environmental | Occlusion Resistant,<br><br>Provides systemic welfare context | Sensor drift over time,<br><br>Lacks behavioral insights | Climate control, air quality, ammonia levels |
| Physiological | Direct stress/disease indicators<br><br>Supports individual health profiling | Invasive,<br><br>Hardware fragility<br><br>Low scalability | Vaccine response, sleep patterns, HRV |

## 3.1 Visual Sensing

Visual sensing technologies remains a cornerstone for precision poultry welfare monitoring, offering non-invasive and scalable solutions to evaluate behavioral and physiological welfare indicators. From thermal imaging to video-based behavior recognition, the various applications of visual systems have significantly expanded in recent years. Figure 6 displays sample RGB, and thermal images captured from laying hen facility, showcasing how visual and thermal cues can reveal subtle welfare indicators such as stress and disease in young hens.

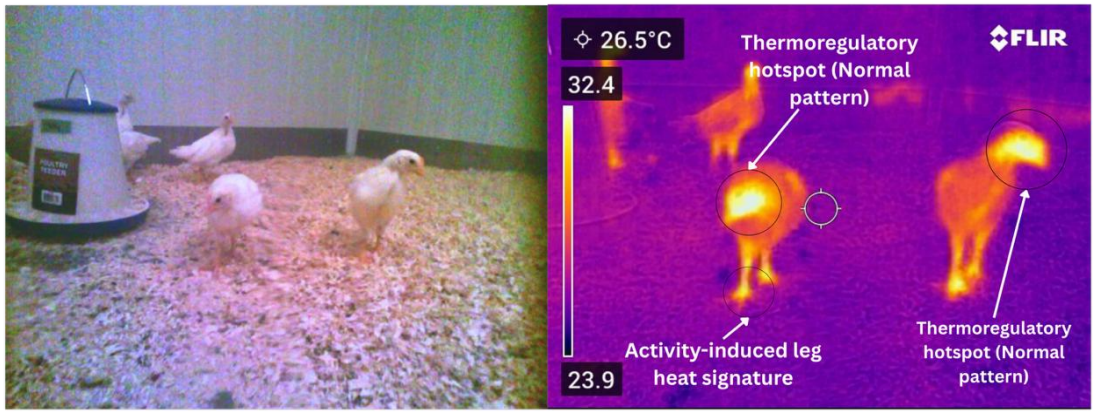

Figure 6. Sample RGB and thermal images from laying hen facility. RGB images support behavior, activity and Gait recognition while thermal imaging detect surface temperature aiding in anomaly detection.

Recent studies have demonstrated the evaluative capabilities and limitations of visual sensing systems in precision poultry farming. Schreiter et al. [62] effectively demonstrated the diagnostic potential of Infrared Thermography (IRT) for plumage assessment, particularly in detecting severe feather damage. However, its reduced sensitivity to moderate feather loss emphasizes a critical limitation in current image resolution and thermal sensitivity thresholds.

Similarly, continuous video recordings coupled with machine learning framework have been utilized to monitor feeding, drinking and activity levels in group-housed broilers [63]. Despite strong performance, the systems struggled to distinguish between active and inactive states, an issue amplified by visual occlusion and ambiguous behavioral definitions in densely stock housing. Guo et al. [64] corroborated this finding, reporting that feeders and overhead equipment obstructed birds in top view images, leading to missed detection in an otherwise high-performing model. Although focused on broilers, the structural and environmental similarities make these findings transferrable to laying hens where similar occlusion and definition ambiguities exist.

These challenges highlight the need for a standardized behavioral taxonomy to reduce misclassification, and the adoption of occlusion aware systems that utilize either depth cameras or multi-angle imaging systems.

Moreover, visual sensing studies are being explored in multimodal context. Putyora et al. [65] combined infrared video and EEG sensors to monitor sleep behavior under various environmental stressors. Although no multimodal sensor fusion model was implemented, the dual-modality approach enabled detection of temperature-induced sleep disturbances, which would likely be overlooked by visual data alone.

In summary, the limitations of visual sensors suggest the need for complementary modalities. Acoustic sensors are suitable complementary sensors with ability to capture welfare relevant data even when visibility in obstructed. Visual systems are utilized mostly for spatial and postural analysis of laying hens and its environment, while acoustic sensors are more suited for acoustic analysis suitable for occlusion and lighting issues. Table 6 showcases a comparative summary of both sensors.

Table 6. Comparison of Visual and Acoustic Sensors for Poultry Welfare Monitoring

| Criteria | Visual Sensors | Acoustic Sensors |
|---|---|---|
| Data types | RGB, Thermal Images, Video | Acoustic Signals / Vocal Patterns |
| Strengths | Behavior Recognition, GAIT analysis, Visual Observation | Detect distress calls, Respiratory Issues, Emotional State Recognition |
| Weaknesses | Occlusion, Poor lighting, Sensor drift | Background noise, Vocal overlap |
| Environmental Sensitivity | Sensitive to lighting, camera angles and image/video quality | Sensitive to environmental noise. |
| Computational Load | High (especially with video) | Moderate |
| Use Cases | Gait, Behavior Recognition, Posture | Cough detection, Stress Vocalization |

## 3.2 Acoustic Sensing

Acoustic sensors offer a robust alternative or complement to visual systems, because they are less susceptible to occlusion and lighting variations. Acoustic sensors offer unique advantages for continuous, non-invasive welfare monitoring, particularly in recognizing nuance, early-stage anomalies in behavior and physiology. Numerous studies utilized machine learning techniques with acoustic features, such as Mel-frequency cepstral coefficients (MFCC) & spectrogram for various tasks like respiratory disease detection and emotional state recognitions [66,67]. Figure 7 illustrates a typical spectrogram, waveform signal graph, and MFCC of chicken vocalization in a poultry farm. The energy waveform plot highlights regions of vocal inactivity (quiet zones) and sharp amplitude spikes (highlighted in red), with few spikes surpassing 0.6. The vocal spikes shown in the waveform chart align with the spectrogram, particularly within 3kHz – 5kHz commonly associated with chicken vocalization. The MFCC visually differ between normal static vocalization and distress calls. There are clear distinctions of more orange colors which represent more spectral activity and stronger harmonics when compared with the purple which shows less spectral richness.

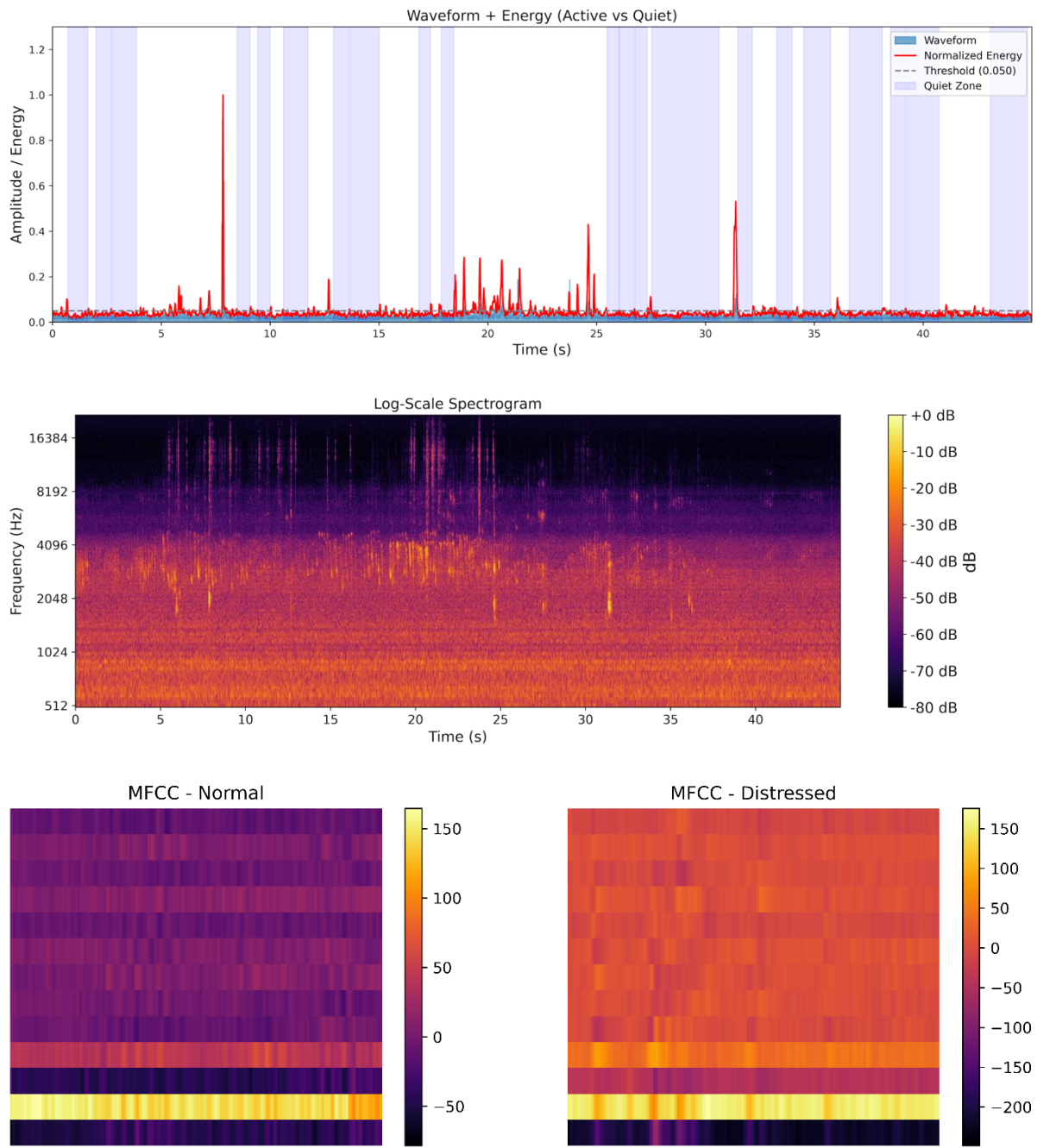

Figure 7. Waveform, spectrogram, and MFCC representations of poultry vocalizations. The waveform shows signal energy over time, while spectrograms and MFCCs extract frequency-based features used in acoustic recognition.

Zhou et al. [67] successfully demonstrated that portable acoustic sensors can detect cough in commercial farm settings, a crucial step towards real world deployment. Ji et al. [68] further reported correlations between vocal patterns and productivity metrics, such as egg weights and laying frequency, highlighting potentials of vocal biomarkers as indirect indicators of physiological performance.

However, despite promising results, most studies face persisting issues with background noise, overlapping vocalizations and acoustic interference. Puswal et al. [69] utilized acoustic sensors to investigate the relationship between acoustic features and morphological parameters but found no significant correlation, possibly due to noise contamination. Similarly, while acoustic data has been utilized for chick sex classification [70], ambient noise limits its broader applicability

To address these challenges, Soster et al. [71] utilized advanced noise suppression algorithms that improved feature clarity in broilers. Similarly, signal segmentation strategies based on Short-Time Zero Crossing Rate (STZ) and Short-Term Energy have been utilized to address noise in poultry acoustic monitoring [66].

These efforts, while promising, showcase a critical gap. The need for scalable, open source and context adaptive noise filtering pipelines that can operate in dynamic farm conditions. Additionally, microphone placements is rarely optimized or considered. Similarly, most acoustic datasets are collected in group-housed settings, complicating the classification of individual chick's vocalization. This highlights the urgent need for localized sensing innovations such as directional or wearable microphones that aid in individual level monitoring

### 3.3 Environmental and Physiological Sensing

Environmental and physiological sensors complement behavioral sensing by providing quantitative insights into ambient conditions and internal states. While environmental sensors are cost-effective and scalable, physiological sensors offer the most granular insights but are invasive and face deployment challenges in commercial farms. These technologies offer a deeper understanding of the internal and external conditions that affect behavior, health and productivity of laying hens.

Zheng et al. [72] utilized intelligent, portable monitoring units to measure temperature, ammonia (NH3), and carbon dioxide (CO2) variations inside a commercial stack-deck cage laying hen house. While their study highlighted seasonal variation in environmental control, its low temporal resolution likely obscured transient but welfare critical spikes in ammonia or heat levels, underscoring the need for higher-frequency data collection. Li et al. [73] further explored environmental influences using multivariate data mining techniques, identifying wind speed and humidity as key drivers of air quality. Though focused on broilers, the analytical approach showcases how advanced statistical modelling can aid in sensor deployment strategies and environmental management in laying hen facilities.

In physiological sensing, Shimmura et al. [74] employed inertial sensors to detect 11 hen behaviors, behavioral frequency, spatial activity, and their transition. This study introduced "behavioral connectivity", a dynamic welfare indicator reflecting the transition probabilities between behaviors. While conceptually robust, the model's diminished accuracy in cage systems suggest a need for calibrations based on housing type.

Kang et al. [75] investigated the sensitivity of physiological sensors in detecting subtle changes post vaccination. The study revealed that activity levels, distance travelled, and feeding behaviors were more reliable welfare indicators after the vaccine than traditional vital signs like cloacal temperatures, feed intake and foraging. This study contradicts conventional assumptions and suggests that behavioral metrics should be considered as a potentially superior indicator for early disease detection. It also highlights the value of feature weighting when creating multimodal AI systems.

Putyora et al. [65] incorporated electroencephalography (EEG) and Infrared cameras to assess sleep quality of hens under thermal stress, revealing that elevated nighttime temperatures significantly suppressed REM and slow-wave sleep. This study not only affirms the importance of multimodal sensing but also reveals a critical blind spot: biochemical markers, such as fecal cortisol, remain largely unexplored, representing an untapped perspective in physiological insight.

Ahmmed et al. [76] introduced long-term ECG monitoring for heart rate variability (HRV) in free moving hens, offering a low power alternative which holds the potential for early detection of circadian rhythms disruption. However, practical issues such as sensor dislodgement due to pecking highlights the engineering challenges of long-term deployment in commercial environment.

These studies showcase environmental and physiological sensors are contrained by technological fragility and contextual variability. The dominating reliability on behavioral indicators rather the physiological indicators after vaccination [75], emphasis the need for more adaptive multimodal fusion strategies which weighs certain modalities over others. Additionally, the recurring omission of biochemical sensing restricts the thorough holistic understand of poultry welfare.

Similarly, maintenance concerns such as calibration drift, pecking damage, and sensor degradation are rarely reported but critically affect system reliability. The development of self-reporting sensor capabilities will be essential to ensure long term data integrity and deployment feasibility in poultry environment.

### 3.4 Comparative Analysis of Sensing Modalities

Table 7 presents a comparative overview of the various sensing technologies deployed in poultry systems, from both broiler and laying hen studies, showcasing a direct comparison across sensor types, target applications, accuracy, and known limitations.

A major challenge across all sensors is long-term measurement reliability. Sensor drift, gradual changes sensor output even when the environmental conditions remain constant, and calibration drift, measurement error due to gradual shift in a sensors value, can significantly erode both sensitivity and accuracy over time. Without routine recalibration and diagnostic checks, long term deployment on commercial farms risk accumulating data inaccuracies, ultimately undermining decision making.

Visual and acoustic sensors demonstrate strong performance in behavioral recognition and disease detection respectively but are often challenged by environmental noise and occlusion. Physiological sensors offer more individualize sensitive welfare metrics, particularly in stress and circadian rhythm assessment but suffer from hardware related issues.

No single modality provides comprehensive coverage across all key welfare dimensions, emphasizing the need for multimodal fusion systems to integrate all the data streams.

Table 7. Summary of Sensing Technologies used in Poultry Systems: Applications, Performances and Limitations

| Sensor Type | Brand / Model / Source | Targeted Application(s) | Efficiency / Accuracy | Results & Limitation | References |
|---|---|---|---|---|---|
| Acoustic | Zoom H4n Pro | Distress detection via vocalization | 95.07 % for chicken distress calls | Efficiently identifies chicken distress calls but model's performance decreased during continuous recoding due to presence of sudden noise. | [77] |
| Acoustic | Lenovo T505 Digital Voice Recorder | Sex of day old chickens via vocalization | Accuracy of convolutional neural networks (CNN), long short-term memory (LSTM), and gate recurrent unit (GRU) was 74.55%, 75.73% and 76.15%, respectively | Fairly worked but there aren't gender differentiating vocal features compared to adult chickens. | [70] |
| Acoustic | Shinco Recorder A01 | Respiratory disease recognition via vocalization | > 90% on all models used (PV-net1, PV-net2 and PV-net3) | Effectively identified cough sounds but presence of environmental noise | [67] |
| Acoustic | Knowles FG-23329 | Differentiate between four distinct vocalizations of broiler chicken (pleasure notes, distress calls, short peeps, and warbles) | Achieved an overall balanced accuracy of 91.1% in vocalization classification | Effectively classified four distinct vocalization but presence of background noise. | [71] |
| Acoustic | Sony PCM-A10 recorder | Analysis of morphological and acoustic | Identified that both male and female chickens differ significantly | Effectively identified morphological | [69] |

| | | features of male and female domestic chickens | | differences but mentioned environmental variations and weather as limitation that impacted acoustic signals | |
|---|---|---|---|---|---|
| Acoustic | Acoustic: Knowles FG-23329 & Elmetron pH meters<br><br>Physiological & Biochemical Analysis: DetectX Cortisol Enzyme Immunoassay Kit, Sysmex-XN-VET impedance analyze, Burker's hematological chamber, Beckman Coulter AU680/AU5800 chemistry analyzers | Identify vocalization patterns of broiler chickens throughout different times of the day and its effect on physiological parameters. | Identified changes in vocalization throughout the day. | Showcased changes in vocalization throughout the day but some issues with annotations | [78] |
| Environment | DHT22 sensor (for temperature and relative humidity)<br><br>MQ-137 electrochemical sensor (for air ammonia content) | Low cost, Hardware, and software instrument to monitor environmental conditions of poultry farms | 90% similarity with proposed monitoring instrument and commercial equipment | Effectively monitoring environmental parameters but its MQ-137 ammonia sensor was most effective at lower concentratio | [79] |
| | LDR (light dependent resistor) sensor (for luminosity) | | | ns and data lost due to unreliable internet connections | |
| Environment | DustTrak II 8533 aerosol monitor<br><br>Korno GT-1000<br><br>Elitech RC-4HC<br><br>Testo 425 | Evaluation of data analytics methods for processing multivariate environmental data in broiler houses during winter conditions. | Spearman and PCA results showed that the in-cage wind speed, aisle wind speed, and relative humidity played critical roles in indoor air quality distribution during broiler rearing | Effectively identified factors that affected indoor air quality but Data was collected from only half of a symmetric broiler house during winter, under restricted ventilation, limiting generalizability across seasons, housing designs, and environmental setups | [73] |
| Visual | Model PRO-1080MSFB | Monitor chicken activity index | YOLOv8 and DeepSORT combination demonstrating the highest performance, achieving a multiobject tracking accuracy (MOTA) of 94% | Effectively tracked activity of poultry but visual occlusion and camera field of view posed as major issues. | [39] |
| Visual | Intel RealSense D415 | Investigate walking characteristics of broilers using automated pose estimation. | Successfully extracted 7 pose features from broiler walking videos using automated pose estimation | Effectively identified Gait characteristics but had issues with low environmental lighting | [80] |
| Visual | Sony HDR-CX405 | Detection and | Accuracy ranged from 0.31 to 0.84 | Successfully classified | [63] |

| | | classification of individual broiler behaviors (feeding, drinking, active, inactive) | across behaviors; feeding and drinking well-detected | feeding and drinking, but struggled with accurately detecting activity vs inactivity. Occlusion, low lighting, and identity tracking issues were major challenges. Training with diverse setups improved robustness, but manual post-processing was still required. | |
|---|---|---|---|---|---|
| Physiological Sensors | MAX30003, Maxim Integrated  CC2642, Texas Instruments | Developed a low-power wearable bioelectric recording system capable of monitoring heart rate and its variability through ECG signals | Demonstrated the feasibility of continuous long-term measurement of heart rate (HR) and heart rate variability (HRV) indices in freely moving chickens using a wearable backpack electrocardiography (ECG) recording system | Limitations include possible electrode damage due to chickens pecking on the device. | [76] |
| Physiological - Wearable (IMU), Visual | MTw2 Awinda wireless motion tracker, Xsens  GoPro Hero 7 | Classification of laying hen behaviors by activity intensity (static, semi-dynamic, highly dynamic) | Overall accuracy ~90%; F1-scores: 0.89 (static), 0.91 (semi-dynamic), 0.87 (highly dynamic); Bagged Trees model performed best | Demonstrated feasibility of wearable IMUs for high-accuracy activity classification. However, small sample size (2 hens) | [40] |

| | | | | and imbalanced datasets limited generalizabil ity. Sparse data for highly dynamic behaviors impacted model robustness. | |
|---|---|---|---|---|---|
| Physiologi cal - Wearable (IMU) | TSDN151 inertial sensor (ATR-Promotions) with 3-axis accel. & gyro | Spatiotemp oral classificatio n and mapping of 11 laying hen behaviors in cages and floor pens | Successfully recognized and spatially mapped 11 behaviors; behavioral diversity higher in floor pens | Enabled detailed spatiotempor al behavior tracking and transition analysis Limitations include reduced classification accuracy in cages, potential mislabeling, and lack of data independenc e. Broader validation across breeds and environment s needed for generalizatio n. | [74] |
| Physiologi cal | FitBark 3-axis accelerometer s, TrackLab UWB system, video cameras | Detection of behavior, activity changes, and clinical signs in laying hens after vaccine challenge | Accurately identified strong indicators such as activity level, depression, feeding, and sitting; weak for temperature/respir atory metrics | Demonstrate d that activity sensors can detect health-related behavioral changes earlier than | [75] |

| | | | | human observation. Limitations include sensor variability, confounding factors (e.g., habituation), and video blind spots. | |
|---|---|---|---|---|---|
| Physiological | MTw2 Awinda Wireless 3DOF Motion Tracker (IMU) from Xsens | Automatically recognize the activity levels of individual laying hens using wearable IMU sensors and deep learning with signal imaging | ResNet50 model achieved an accuracy of 0.9991, Shallow model achieved an accuracy of 0.9998 | Limitations include imbalanced dataset and study focused on individual laying hens. | [81] |

## 3.5 Real-World Deployment Challenges in Commercial Farms

Translating multimodal AI and sensor technology from lab environments to commercial farms remain a challenge. These limitations are not merely technical but systemic, stemming from complex interplay of environmental variability, sensor fragility and lack of standardized behavioral taxonomies.

Commercial poultry farms are inherently dynamic, characterized by temperature fluctuations, dust, high ammonia concentration, lighting variations and waste build-up, all which interfere with sensor signal quality and hardware longevity. For example, Yang et al. [39] highlighted that dust and debris frequently interfere with both acoustic and visual sensors leading to inaccuracies during long term monitoring of poultry behavior and activities. Similarly, Fodor et al. [80] observed that natural lighting inconsistencies required extensive post processing to normalize video-based data, highlighting the vulnerability in vision-based models to environmental lighting conditions. Unlike controlled research facilities, commercial farms operate with non-stationary environmental conditions that cause data drifts, reducing sensor and model reliability.

Behavioral annotations in real world poultry settings remain ambiguous and context dependent, especially for subtle welfare indicators like low amplitude vocalizations or actions with similar motions (e.g. aggressive pecking vs feeding or comfort preening vs feather pecking). Distinguishing and annotating these behaviors requires careful observation such as intent, target, duration or context of interaction. Guo et al. [82] encountered similar challenges while conducting video-based behavior monitoring for broiler chicken at different ages. Issues with misclassification when the chickens were near feeders and drinkers, yet not drinking, and misclassifications in certain postures. The absence of standard ontology for poultry behavior makes training robust AI models difficult.

Moreover, sensor deployment presents additional logistical challenges. While wearable sensors offer direct physiological measurements, they are impractical in high-density commercial farms due to their impact on bird behavior, risk of injury, and maintenance burden. Noninvasive sensors (e.g. overhead cameras, acoustic arrays, and environmental probes) offer more viable solution. However, these solutions remain vulnerable to degradation from environmental exposure.

Ruggedized equipment such as Wildlife Acoustics' Song Meters may withstand harsh conditions but are economically prohibitive for large scale poultry farm which typically run on narrow profit margins. This economic reality creates a disconnect between technical feasibility and practical deployment warranting the development of cost-effective, yet durable sensors optimized for the unique constraints of poultry production system.

Additionally, model robustness under real-world conditions remains under addressed in literature, particularly in sensor failure, occlusion and unstructured behavioral patterns. Table 8 presents a comparative cost-benefit analysis of selected AI-based poultry welfare monitoring systems highlighting the tradeoffs between initial investments, sensor sophistication and expected benefits, emphasizing the economic challenges that limit practical deployment despite technical feasibility.

Table 8. Estimated Costs, Operational Demands, and Anticipated Benefits of AI and Sensor Systems for Laying Hen Welfare

| Technology | Estimated Upfront Cost | Annual Operating Cost (Relative) | Key benefits | Expected Return on Investment (Qualitative) | Notes | References |
|---|---|---|---|---|---|---|
| AI-Powered Vision + Acoustic Monitoring System | High (multiple cameras & mics) | Medium (Storage, Software Subscriptions & Maintenance) | Early detection of diseases such as respiratory issues via acoustic data, behavioral changes like lethargy<br><br>Automated Bird Counting and Weight estimation | Moderate – High:<br><br>Increase in production efficiency due to reduced mortality and improved welfare<br><br>Potential saving from early disease interventions | Cost varies significantly with brands, resolution, AI sophistication and coverage area.<br><br>Also requires internet connectivity and computing power. | [83]<br><br>Industry Report: [84] |

| | | | Early detection of abnormal behaviors<br><br>Reduced Labor for Manual checks<br><br>Automated Detection of dead birds<br><br>Automated detection of eggs | | Infrastructure to setup sensor would potentially increase costs. | |
|---|---|---|---|---|---|---|
| Multimodal Environmental Control (Temperature, Light, Humidity, Ammonia, CO2) | Medium (depending on sensor types, control panel complexity, and number of zones covered) | Low - Medium (Sensor calibrations, Software updates, Maintenances) | Automatic optimization of ventilation, heating and cooling based on real time data.<br><br>Prevention of harmful gases (ammonia, CO2)<br><br>Reduced Energy Consumption<br><br>Improved air quality | High:<br><br>Immediate ROI from reduction in energy cost<br><br>Increase growth rate and egg production due to optimal conditions | ROI often reflect immediately due to energy saving and improved bird productivity.<br><br>Includes AI modules to dynamically change environmental conditions. | [85,86] |
| Thermal Imaging System for Health and Density Monitoring | Low - Medium | Low (Software, Maintenance) | Automatic detection of sick or stressed birds via | Medium:<br><br>Improved welfare and mortality | Can be combined with vision systems for robust | - |

| | | | heat signature | | multimodal approach. | |
|---|---|---|---|---|---|---|
| | | | Monitoring of housing density and huddling patterns. Identification of cold spots in poultry house, allowing for precise environment adjustments | | | |
| Wearable Sensors for Individual bird Monitoring | High | High (Data transmission, Batteries, Maintenance) | Real time health monitoring<br><br>Detection of physical ailments like lameness, abnormal activity or stress<br><br>Precise tracking of growth and feeding behavior Enables individual intervention. | Low:<br><br>Highly speculative ROI due to high cost per bird, coupled with large cost of maintenance.<br><br>Benefits are in research facilities or high value poultry breeds | Impractical for large commercial farms due to cost and deployment complexities<br><br>More prevalent in research<br><br>Challenges with battery life and invasive nature of sensors | [87] |
| Multimodal AI platform for Farm wide Data Analytics | Very High (cloud storage, advanced analytics, | High (Platform subscriptions, Data analysis, IT support, | Holistic view of poultry farm operations | High (Long term):<br><br>Drastic improvement in | Requires large initial investments.<br><br>Gained value from | Industry Report: [88,89] |

| | web app hosting) | Maintenance) | Optimized resource management<br><br>Improved decision making for profitability and sustainability<br><br>Predictive capability for disease outbreaks, feed consumptions & market trends | operational efficiency<br><br>Potential long-term ROI despite large initial cost.<br><br>Enhanced overall farm management<br><br>Reduced Mortality Improved feed conversion rations across flock cycles | combining multiple data sources to generate insights.<br><br>Web app systems would have interactive dashboard with customized alerts. | |
|---|---|---|---|---|---|---|

Quantitative ROI data in AI based poultry systems are limited, especially for multimodal systems. As such, ROI estimates are qualitative and based on insights from academic studies, industry reports and case study evaluations.

### 3.6 Implications of Housing Systems Design of Sensor Performance and Placement

### 3.6.1 Aviary Systems

Aviary housing systems are cage-free housing systems that has a multi-tiered, vertical structured designs with ramps and perches that allow bird to freely move and display natural behavior. This housing system allows for high density housing while allowing freedom to express natural behaviors, however it has limitations such as visibility restrictions, noise and behavioral variations.

In vision-based sensors, Lamping et al. [90] highlighted that image quality in aviary housing was often reduced, leading to lower model accuracy of neural network outputs. Similarly, YOLO-based models can detect hens on the floor with high accuracy, but dust-bathing behavior remains a challenge due to inconsistent lighting and movement complexity [91]

Although RFID systems effectively track bird location and movement, they fall short in linking this information with crucial welfare indicators such as feed intake or stress. Welch et al. [92] and Sibanda et al. [93] noted that these systems cannot recognize shifts in health status in real-time because welfare data is analyzed only at post-production, creating latency issues that reduces predictive accuracy.

Wearable Inertial sensors, though successful in behavioral classification in small, controlled farmers, fail to generalize in aviaries of larger sizes [40]. Additionally, sensor fusion in aviary housing systems must handle dropout-prone data types and asynchronous behavioral patterns, which complicate early fusion.

### 3.6.2 Caged Systems

In contrast, caged housing systems are more controlled and confined, which reduces the effect of environmental influences on sensors. The standardized, stacked layout of cages allows for a structured sensor placement, especially vision-based systems, but the benefits are often nullified by physical obstructions and restricted movement.

Due to cage infrastructure (e.g. metal wires, feeders and drinking stations), camera occlusion is likely. Wu et al. [94] , in dead chicken detection, solved this by collecting data during feeding times to encourage bird movement, improving visibility for deceased birds. Li et al. [95] further improved beak abnormality detection by incorporating attention and convolutional modules to reduce information and improve feature extraction but still mentioned persistent obstruction from cage mesh and lighting shadows. Timed data collection serves as a potential solution for visual data collection, while the incorporation of attention and convolutional layers improve detection accuracy of visual-based models

Similarly, acoustic sensing face challenges in caged housing. Zhou et al. [67] noted that microphone had difficulty isolating relevant signals due to ambient noises such as background ventilation and feeding noises, while portable microphone, though more precise, are prone to damage from pecking and bird interference.

In commercial laying hen farms, thermal and depth cameras were used to overcome the occlusion and lighting challenges. However, Luo et al. [96] reported black noise points in NIR and depth image. Despite these limitations, caged systems benefit from environmental certainty and reduced behavioral variability making it ideal for late fusion strategy, where sensor redundancy can reduce ambiguities. However, there are ethical concerns regarding restricted movement and overcrowding.

### 3.6.3 Floor-based Systems

Floor-based poultry systems provide chickens with more freedom to move and express natural behaviors, however, due to challenges with bird density, overlapping movement and environmental variability, they need sensors that are robust to high motion variability, occlusion and volatile surface visibility.

Computer-vision based systems have showcased high accuracy for behaviors in floor-based housing. For instance, YOLO-based models such as YOLOv5x-hens and YOLOv8x-DB achieved 93% accuracy in detecting litter floor behaviors, even under low lighting [97,98]. However, high stocking density, low camera angles, and dust accumulation degrades model's performance, which environmental cleanliness and camera placement very vital.

The utilization of RFID and wearable inertial sensors allows individual level behavioral monitoring, but generalization remains a concern. Derakhshani et al. [40] and Welch et al. [92] showcased that while these sensors accurately predicted disease outcomes and classified activity levels, performance accuracy was based on infrastructural designs such as vertical location of feeders, available space of movement, which varies across barns. This raises concerns about generalization and scalability of sensor fusion models as barns and facility vary.

The fusion of multiple modalities, such as audio, thermal imaging and depth sensing, could potentially enhance effectiveness of floor monitoring. For instance, multimodal imaging approaches outperformed standard 2D detection methods in detecting feather damage and bumblefoot [98,99]. Similarly, acoustic based models, such as CNN-MFCC framework, achieved over 98% accuracy in detecting distress calls, respiratory diseases and stress related vocal changes [100]. These findings highlight the need for sensors that are complementary in fusion architectures (vision handles spatial details, thermal handles surface anomalies and audio detect welfare indicators that other modalities can't recognize).

In contrast to caged and aviary systems, floor systems face hybrid occlusions due to overcrowding and equipment obstruction, which require adaptive camera calibrations and placement with dynamic fusion models that weigh sensor confidence based on real-time signal integrity. Future research must also focus on transfer learning techniques which allow for multimodal AI models to work across various barns with different layout.

## 4. Applications in Laying Hen Welfare and Productivity

This section synthesizes current and potential applications of multimodal AI within the poultry farming industry, with a focus on laying hen systems. There are very limited studies on multimodal data fusion for poultry, especially laying hens but we can utilize lessons from broader livestock domains to provide valuable foundation for enhancing poultry welfare, disease detection, nutrition optimization, and resource management.

### 4.1 Enhancing Welfare and Health Monitoring through Sensor Fusion

The welfare of laying hens directly affects productivity, mortality, and farm sustainability. Suboptimal welfare conditions manifest as elevated stress levels, behavioral abnormalities, increased disease outbreaks, and decreased egg production.

Traditional unimodal monitoring systems lack context and reliability to detect subtle changes in health and behavior. In contrast, multimodal AI systems can fuse different modalities such as chicken vocalization, video-based detection of behavior, physiological data, and thermal imaging, to provide a transformative effect.

### 4.1.1 Multimodal Surveillance and Early Disease Detection in Laying Hens

Although poultry-specific multimodal studies are limited, fusion approaches have shown potential in other livestock species. For example, Dhaliwal et al. [45] combined facial biometrics and accelerometer data using DenseNet-LSTM model, achieving 99.55% accuracy in lameness detection in dairy cows. Similarly, Chae et al. [101] employed feature-level fusion to implement an audio-visual system to localize coughing pigs using deep audio stream analysis combined with object tracking.

However, applying similar strategies to poultry present unique challenges. Chicken vocalizations are more ambiguous, rapid and context-dependent that those of pigs or cows, complicating interpretations. Additionally, visual occlusion from flock density can also hinder surveillance. These differences demonstrate the necessity for poultry specific adaptation of multimodal systems.

These species-specific differences emphasize the need for tailored multimodal systems for poultry, On thermal imaging front, Sadeghi et al. [32] implemented an early detection system by combining thermal imaging with machine learning to classify Avian influenza and Newcastle Disease with up to 100% accuracy within 24 hours of infection. Incorporation of Dempster-Shafer fusion further improved reliability when primary classifier underperformed.

However, this study is limited by small dataset and controlled lab conditions prevent real world generalizability. This emphasizes a need for vast datasets and model adaptation to dynamic farm settings with variable lighting, occlusion and equipment interference.

Sound-based systems, such as Zhou et al. [67] and Hassan et al. [102], showcase the potential for early disease detection but also faced challenges such as overlapping vocalization, barn noise and echo effects. In poultry, distress calls aren't as distinct as pigs or cows, making classification more difficult without additional context from videos or physiological metrices.

There is a significant gap towards the deployment of these models at scale under the complex, noisy and occlusion-heavy nature of commercial poultry farms. Future studies should consider incorporating multimodal fusion to supplements in areas where one modality is lacking.

### 4.1.2 Emotional State Recognition and Behavioral Detection in Poultry

Emotional and Behavioral state are essential welfare indicators, and their assessment must be grounded in poultry specific behavior ethograms. We propose the inclusion of a standardized ethogram (Table 9), categorizing key visual, acoustic, and physiological indicators of welfare related states such as aggression, fear, comfort, and illness

Table 9. Standardized Ethogram for Laying Hens Across Visual, Acoustic, and Physiological Modalities

| Behavior Name | Visual Indicators | Acoustic Indicators | Physiological Indicators | Reference |
|---|---|---|---|---|
| Feeding / Drinking | Beak directed toward feeder/drinkers, <br><br>Repetitive pecking, <br><br>Movement towards feeder, <br><br>Frequent feeder visits | Distinctive pecking sounds (~1-3 kHz), <br><br>Increased vocalization before feedings | Measurable water and feed intake | [71,97,100,103] |
| Dustbathing | Ventral or Side-lying posture, <br><br>Wing-shaking, <br><br>Substrate raking with beak/legs | Typically silent, though low clucking may occur | Elevated Corticosterone levels when dustbathing substrate is absent <br><br>Thermal imaging show increased peripheral circulation | [40,91,103,104] |
| Aggression | Rapid pecking | Sharp squawks and | Elevated Cortisol, | [105] |

| | | high-energy calls, <br><br> Increased flock noise | Wounds/Injury <br><br> Deteriorated, <br><br> Feather condition | |
|---|---|---|---|---|
| Piling / Overcrowding | Formation of dense clusters, <br><br> Overlapping bodies, <br><br> Restricted movement, <br><br> Often near walls or corners | Silent or distressed vocals during smothering events | Elevated core temperature, <br><br> Heat stress, <br><br> Asphyxiation risk. | [106] |
| Nesting / Egg laying | Entry and Sitting in Nest Boxes <br><br> Mislaid eggs on litter floor <br><br> Nesting site selection <br> Nest Guarding | Quiet or Broody Clucking <br><br> Low Vocal Activity | Egg production rates <br><br> Quality of egg (egg shells) | [93,107,108] |
| Exploration | Movement around surrounding, <br><br> Exploratory pecking at objects | Mostly silent, but may have occasional murmurs | Quantifiable activity levels | [40,93] |
| Comfort Behaviors | Preening, <br><br> Stretching, <br><br> Resting, <br><br> Perching, <br><br> Excellent Plumage Condition | Low Frequency Murmurs, <br><br> Repetitive, Harmonic Warbles | Optimal Body Temperature <br><br> Stable Respiration | [40,71,103] |
| Fear / Distress | Freezing, <br><br> Piling, <br><br> Crouching Posture, <br><br> Escape Attempts | Repetitive High-pitch calls, <br><br> Vocal Entropy Increases, <br><br> Alarm and Frustration | Elevated Heart Rate, <br><br> Increase Cortisol levels, <br><br> Drop in peripheral Temperature, | [67,71,100,103,106] |

| | | calls mostly <5kHz | Increased Immune Suppression | |
|---|---|---|---|---|
| Disease | Lethargy, Abnormal Posture, Keel Bone Damage, Increased Resting, Uncoordinated Movement, Bumblefoot, Increased preening/dustbathing due to ectoparasites, Abnormal Beak features Plumage Damage | Sneezing, Abnormal Vocalization, Rales (Poor breathing) | Elevated Surface Temperature (Fever), Irregular Egg Production, Presence of Parasites (e.g. Ascaridia galli, cestodes) | [32,95,98,102,103] |
| Plumage Condition | Feather Loss Patterns | No Known Acoustic Indicators | Increased Heat Loss and Feed Consumption Detectable thermal differences | [99] |

Emotional recognition allows for a more nuanced welfare assessment of the inner workings of poultry. Although studies in cattle, such as using WHISPER audio features to categorize affective vocalizations [109] , has shown potential, these affective states are more complicated to identify in poultry due to less structured vocal patterns. Commercial applications are beginning to emerge, like CluckTalk and Cluckify, developed by MooAnalytica, which aims to decode chicken vocalization via mobile platforms.

Pose estimation frameworks like ViTPose, GroundingDINO have been validated in larger livestock [110] and are being adapted for poultry. Practical applications have already emerged with Fang et al. [111] which successfully estimated the poses of multiple chickens using 2D deep learning techniques based on transfer learning techniques. Further refinements by Fang et al. [112] addressed distinct posture problems (standing and lying) in caged environments by leveraging 3D depth information alongside color images and customized a computer vision algorithm to address issues of lighting conditions. More studies have been conducted towards behavior detection in poultry with innovations like FCBD-DETR (Faster Chicken Behavior Detection Transformer developed by Qi et al. [113] to enhance detection speed, accuracy and generalization in chicken behavior analysis.

In pose estimation and behavior tracking in poultry, significant opportunities for innovation remain, particularly through the fusion of multiple modalities. Combining pose estimation with video-based behavior detection and acoustic analysis holds the ability to identify subtle, early indicators of welfare issues. For example, the early detection of lameness or disease, which often shows up in subtle changes in stride length or weight distribution, could be significantly improved. Similarly, the identification of bullying and avoidance behavior could be identified by analyzing the relative distance between birds and precise body orientations, which can combine with aggression related vocalizations. This integrated approach shows a more comprehensive and accurate assessment of laying hen welfare.

Despite considerable progress, most studies still utilize supervised classification, where well labeled behavioral data is required. This dependency is problematic, where labelling is resource intensive and observer agreement is low.

To mitigate these limitations, unsupervised or semi-supervised clustering methods like HDBSCAN or Deep Embedded Clustering, should be heavily explored as potential solutions to identify latent behavioral states. These techniques group similar behavioral episodes (e.g. pecking, pacing, freezing), without prior labelling.

## 4.2 Multimodal AI for Precision Nutrition in Layers

Traditional feeding systems in laying hens prioritize for uniformity, neglecting the individual variations, metabolic demands, and productivity. In contrast, AI-based precision feeding leverages sensor data to curate nutritional requirements dynamically. Multimodal systems bring value by contextualizing feeding behavior with the environment, behavior and health indicators.

### 4.2.1 Monitoring Feeding Behavior and Welfare

Monitoring feeding behavior is central for optimizing nutrition and maintaining production efficiency. Although recent studies are largely focused on single modality (e.g feeding schedules or weight scales) which lack granularity, advances have introduced AI powered methods for monitoring individual bird feeding patterns.

Seber et al. [114] developed a smart feeding unit prototype for broilers, that continuously measured pecking force and feed intake per peck using real time signal acquisition and processing. Findings showcased an average pecking force of 1.39N and feed intake per peck of 0.13g based on 173 pecker per minute. This study serves as a strong foundation for integrating mechanical force sensors with computer vision to monitor feeding behaviors and intake dynamics. Although the system was designed for broilers, it could be adapted for laying hens. However, this study tested on a small sample size (n=7), and lacked environmental variability, limiting its generalizability. Scalability, robustness to barn noise and integration with other modalities such as temperature and weight should be explored further.

During feeding process, the beak-feeder collision produces distinctive acoustic signals. Based on this, Amirivojdan et al. [115] introduced a low-cost, audio-based feed intake estimation system using piezoelectric sensors and a deep learning model. They utilized audio data and trained a VGG-16-based CNN to classify feed pecking vs non-pecking events. The model achieved 92% accuracy and estimated daily feed intakes with a difference of ±8%.

In contrast to force sensors, acoustic approaches are more scalable, lower cost and easier to retrofit into existing commercial farms. For laying hens, that often vocalize during feeding or display social pecking behaviors, additional processing (e.g. noise filtering) would be required to distinguish purposeful feeding from non-nutritive pecking or stress vocalizations.

Both studies showcase sensor driven feed monitoring systems, yet neither utilized multimodal fusion. However, there is a transition from the traditional weight-based feeding to sensor driven monitoring. More studies should try incorporating pecking force with video, and acoustic signals for a comprehensive feed behavior analysis.

### 4.2.2 AI Enhanced Feed Formulation and Predictive Intake Modelling

Multimodal AI can also enhance feed formulation through predictive modeling, incorporating feed composition, environmental variables and bird behavior.

You et al. [116] utilized machine learning regression models to predict the Pellet Durability Index (PDI) using over 2,400 real world feed mill observations. Their results showcased wheat inclusion levels and ambient temperature as key factors that significantly affect pellet quality. This highlights the importance of contextualizing formulation data with environmental inputs to optimize feed consistency for laying hens.

Similarly, Liu et al. [117] combined near-infrared (NIR) spectroscopy and ML classifiers (SVM, RF) to detect microplastics in chicken feed with over 85% accuracy. Although promising, spectral overlap between contaminants limited classification precision, showcasing a challenge in real world implementation.

These studies showcase different approaches to improve feed quality control using sensor integrated systems. Yet most studies remain unimodal and don't factor in the actual hen performance when these systems are implemented. There is a pressing need to develop multimodal systems that fuse feed composition, environmental conditions, and intake behavior into a closed loop nutritional framework

### 4.3 Cross-Species Transferability and Avian-Specific Considerations

It is important to explicitly acknowledge the limitations present when transferring findings from other species to laying hen. The limited availability of multimodal AI studies focused on laying hens require the utilization of analogues from other livestock fields to demonstrate the benefits, technical maturity and proven frameworks of multimodal AI. These cross-species examples provide information of system architectures and inspire translation research but are not intended for direct implementation in poultry systems.

Laying hens have different physiological and ethological characteristics that restrict direct transferability. They possess complex flocking behaviors, high stress vocalization, vertical space utilization (e.g perching) and feather specific behaviors such as preening and feather pecking. Additionally, their physical environments and housing systems introduce unique challenges. Tiered housing systems increase occlusion and reduce visibility for visual sensors, and complicate individual tracking. Similarly, multiple vocalizations in dense housed flocks introduce acoustic interference, making it difficult to detect individual distress calls.

Laying hens, in contrast to cattle and pigs, don't tolerate wearable sensors. They often peck and dislodge at physiological sensors, compromising data quality and longevity of sensors [74] Physiologically, avian biomarkers, thermoregulations and anatomy differ from other livestock animals (Table 1). Therefore, multimodal systems should adapt to specie specific biomarkers to prevent diagnostic inaccuracies.

These differences highlight the need for avian-specific multimodal AI systems, behavioral taxonomies, and data fusion optimizations to carter for the environmental, physiological and behavioral differences of laying hens. Although cattle and swine studies offer valuable insights, the translation to laying hens require further research and on farm validation studies.

### 4.4 Ethical Implications of Continuous Monitoring in Poultry Welfare

As precision poultry farming increasingly rely on sensor-driven monitoring systems, ethical considerations must evolve alongside technological progress. Although multimodal AI provide deeper insights into laying hen welfare, continuous surveillance introduces new ethical implications that require systematic analysis. These include behavioral disruption, autonomy and unintentional stress.

Historically, poultry welfare was assessed using the Five Freedoms framework [1], which focus on hunger, discomfort, pain, fear and freedom to express natural behaviors. However, modern approaches such as the Five Domains Model proved a more nuanced lens, capturing both physical and affective states [3]. Another emerging perspective called Animal Agency recognized birds as sentient beings capable of exerting control and making choices. Welfare shouldn't only be about protecting the birds from harm but enabling them to make meaning choices such as nesting, foraging or dustbathing.

If not thoughtfully designed, continuous monitoring systems risk breaking these welfare principles. Persistent video surveillance may hinder their exploratory behaviors, while intrusive physiological sensors may add stress or disrupt natural activity cycles. Additionally, excessive individual tracking in densely stocked houses could reduce social cohesion. To incorporate innovation with ethics, future AI systems should utilize "ethics-by-design" principles, which include prioritizing non-invasive, flock-level sensing to mitigate behavioral disruption, enforcing transparency and consent protocols where data usage correspond with stakeholder values and developing systems that respect behavioral privacy and avoid enforcing productivity metrics at the expense of the bird's emotional welfare.

### 5. Performance Evaluation and Benchmarking

### 5.1 Summary Table of AI Model Performance

This subsection consolidates performance metrics from recent studies which applied AI & ML for key welfare tasks, including disease detections, behavior monitoring, environmental monitoring, feed optimization etc. Table 9 compares AI model performances of previous studies based on their input modalities, approach, evaluation metrics, specie focus.

Although most studies showcase high accuracy, they are often in controlled research facilities, highlighting the need to validate them in commercial environments, underlining the importance of contextual benchmarking for real-world adoption.

To support this, two additional evaluation dimensions are proposed: the Deployment-Readiness Index (DRI) and the Domain Transfer Score (DTS). These serve as standardized, qualitative benchmarks to support performance metrics with operational and generalization insights. Definitions and scoring criteria are explained in Section 5.3. Notably, the DTS scores in Table 10 showcases that basically no study has "high" score, indicating that most AI models lack robust transferability to other farm setups or production systems.

Table 10. AI Performance in Key Poultry Welfare Tasks

| Study | Application | Data Modality | Sample Size | Model/ Algorithm | Performance Metric(s) | Specie Focus | Key Limitations | DRI | DTS |
|---|---|---|---|---|---|---|---|---|---|
| | | | | | | | | | |

| [80] | Lameness scoring | Video | A total of 109 birds, of 14-days old | DeepLabCut | 2.12px train, 4.83px test pixel error; 0.27 kappa | Broilers | Poor gait birds, limited data | M | L |
|---|---|---|---|---|---|---|---|---|---|
| [118] | Interindividual distance/orientation | Video | 8 distinct stocking densities, each with group of 3 to 10 Jingfen laying hens | Deep learning classifier, YOLOv5m | 99.5% accuracy, 90.9% precision, 93.2% recall, 92.0% F1 | Laying hens | Head/tail differentiation, single replication | H | L |
| [119] | Perching behavior detection | Video | 200 Hy-line W-36 birds | YOLOv8x-PB, YOLOv7, YOLOv8s | 94.8% precision, 95.1% recall, 97.6% mAP | Laying hens | Aging/bone quality, camera cleaning | M | L |
| [117] | Microplastic detection in feed | NIR spectroscopy | 320 chicken feed samples (80 non-contaminated, 240 contaminated ) | SPA-SVM, PLSDA, BPNN, RF | 96.26% accuracy, 100% precision/recall/F1 (PVC) | Chicken feed | Not universally applicable | M | L |
| [81] | Hen activity recognition | Wearable IMU sensors | 2 laying hens | DCNNs (LeNet5, ResNet 50) | 100% accuracy, 1.0 recall, 1.0 precision | Laying hens | Cumbersome sensors, limited subjects | M | L |
| [115] | Feed consumption monitoring | Audio | 10 male Ross 708 broiler chickens | CNN (modified VGG16) | 94% accuracy, 93% F1, 0.97 AUC, 8.7% error | Broiler chickens | Overlapping pecks, feed level | H | L |
| [98] | Bumblefoot detection | Image data | Dataset of 2,200 labeled images of hen feet captured | YOLOv5m-BFD | 93.7% precision, 84.6% recall, 90.9% mAP@0.50 | Laying hens | Batch size/epochs impact | H | L |

| | | | from 720 Hy-Line W36 Laying Hens | | | | | | |
|---|---|---|---|---|---|---|---|---|---|
| [104] | Dustbathing behavior detection | Image data | 6000 images gotten from 800 Hy-Line W-26 Laying Hens | YOLOv8x-DB | 93.4% precision, 91.20% recall, 93.70% mAP@0.50 | Laying hens | Equipment occlusion, dust, camera cleaning | M | L |
| [103] | Behavior, disease prediction | Accelerometer data | 99,998 clean records of accelerometer data gotten from 24 unique poultry chickens | Optimized SMOTE-DT | 97.84% accuracy, 1 G-value, 90% precision | Poultry | Imbalanced data | M | L |
| [93] | Hen sub-population identification | RFID data | 7244 Lohmann Brown laying hens | K-means, agglomerative | 0.7794 Kappa coefficient | Laying hens | Uneven resource load; feed intake not quantified | L | L |
| [120] | H9N2 influenza status prediction | Production data | 3 months of daily laying rates and mortalities collected from three H9N2-infected laying | XGBoost classification | >90% accuracy, >90% recall, >0.85 AUC | Laying hens | Attribute noise from excess information | H | L |

| | | | hen houses | | | | | | |
|---|---|---|---|---|---|---|---|---|---|

## 5.2 Poultry-Specific vs. Cross-Species Results

Although the focus of this review is on laying hen welfare, major insights can be drawn from validated applications in other poultry species and across different animal and human domains. Table 11 compares the performance and design of AI systems used in poultry against livestock, healthcare and other domains, while highlighting transferrable insights, methodologies, performance benchmarks, and the limitations of cross-species generalization.

Table 11. Comparative Multimodal AI Studies from Non-Poultry Domains and Their Implications for Poultry

| Reference | Application Domain | Multimodal Fusion Strategy | Data Modality Used | Model | Evaluation Metric | Key Takeaway for Poultry |
|---|---|---|---|---|---|---|
| [121] | Human Activity Recognition | Early, multi-sensor data fusion | Accelerometer, GPS, light, Profile | LSTM-based neural network | Average F1-score (0.79) | Similar setup can be adapted for hen activity recognition |
| [122] | Skin Disease Classification | Feature-level, multi-attention fusion | Clinical images, dermoscopic images, text | EfficientViT, BERT | Accuracy (79.14%) | Dynamic Weight Loss allocation utilized in this paper enables precise automatic detection of early welfare indicators  Cross attention fusion integrates diverse data streams like visual assessment of plumage and behavioral data seamlessly |
| [123] | Child Emotion | Feature-level fusion | Physiological (BVP, | MLP classifier | Accuracy (74.96%) | Multimodal framework |
| | Recognition | (physiological, facial) | EDA, ST), facial expressions | | | utilized here can be adapted for emotion recognition in poultry |
| [124] | Human Activity Recognition | Feature-level, Late fusion | Videos, inertial data | Two-stream ConvNet, LSTM, Temporal CNN | Accuracy (85.47%) | Fusion strategy can be adapted for poultry behavior recognition. |
| [42] | Personality Traits Recognition | Feature-level fusion (emotion-guided) | Visual, audio, text features | X3D, ResNet-34, SCL, Auto-fusion | Improved Accuracy | The proposed emotion guided fusion and contrastive learning framework offers potential solutions to extracting deeper, more distinguishable features across different modalities addressing the issue of recognizing complex poultry welfare indicators. |

## 5.3 Practical Benchmarking: Domain Transfer and Deployment Readiness

Most studies use traditional performance metrics such as accuracy, precision, or F1-score when comparing AI models, but they often fail to showcase how the model performs in real-world poultry farm conditions. To address this, we proposed two complementary evaluation metrics: Domain Transfer Score (DTS) and Deployment-Readiness Index (DRI). The qualitative measures aim to capture model generalizability and operational feasibility, providing a more holistic view of system performance in real world settings.

The scoring metrics, DTS and DRI scores used in Table 9 were initially developed by the lead author and independently reviewed by a post-doctoral researcher and the supervisory author as part of the inter-rater reliability process. After the review, only two out of the full set of inter-rater ratings differed from the lead author's evaluation. These differences prompted a collaborative discussion between the lead author and the postdoctoral student. This iterative double-review process allowed for scoring consistency, and enhanced reliability of the final framework.

For further details on the derivation and rationale behind the DTS and DRI scoring framework, see Appendix B.

### 5.3.1 Domain Transfer Score (DTS)

The Domain Transfer Score evaluates how robust the model is when utilized to data from various farms, housing conditions, seasons, or animal breeds. It highlights the model's ability to generalize beyond its original training environment, which is critical for commercial scale adoption.

DTS is scored as High (H), Medium (M), or Low (L) based on Dataset Diversity, External Validation & Replication Strategy. If the model meets all 3 criteria, it rates high and 1 or none gets a low score. For example, models trained and tested using data from the same environment, with limited diversity, would receive a Low DTS, while those with extensive external validation would receive a High DTS.

### 5.3.2 Deployment-Readiness Index (DRI)

The Deployment-Readiness Index (DRI) assesses the deployment feasibility of the AI model in commercial farm environments, especially under the infrastructure constraints in poultry production. It considers hardware demands, latency, integration complexity, and the nature of the evaluation environment.

DRI is scored as High (H), Medium (M), or Low (L) based on Model complexity, Hardware requirements, Inference speed, Deployment context and Ease of integration to farm setup. For example, a lightweight model tested in a commercial farm with edge compatible hardware and low latency score High, while a large, cloud-based model validated only using controlled labs would mark Low.

It is important to note that limited studies report key benchmarking parameters like inference latency, model size, or replication strategy. However, the absence of these metrics does not reduce the need for structured evaluation like DTS and DRI. In cases where these parameters are absent, models are scored conservatively using the available information provided.

In conclusion, frameworks like DTS and DRI offer a more holistic and actionable basis for comparing AI systems across both research and deployment contexts.

### 5.4 Evaluation Gaps and Generalizability Issues

Despite the promising evaluation metrics in poultry focused AI system, significant evaluation and generalization gaps remain, especially when models are translated into practical farm deployments. Five major bottlenecks are identified that currently limit the reliability, scalability and adoption of multimodal technologies in commercial poultry farms.

### 5.4.1 Overreliance on Controlled Environments

Although reported precision and recall scores surpassed 90%, these results are derived from highly controlled experimental setups which lack real-world environmental variability. For instance, models for vocal stress detection [100] and perching behavior [119] were trained and tested in acoustically insulated or minimally obstructed houses. This creates a performance mirage that doesn't translate to real-world conditions involving noise, flocking density, ambient noise and dust interference. The lack of domain transfer validation makes these models brittle in new settings. Incorporating Domain Transfer Score (DTS) as part of performance reporting can standardize assessment of generalizability.

### 5.4.2 Inconsistencies in Evaluation Metrics and Benchmarking

Comparative benchmarking across studies is hindered by the absence of standardized evaluation protocols, including unreported latency and inconsistent metric usages (e.g AUC, mAP, F1-Score). For example, some studies in Table 8 utilized mAP to showcase performance while others relied solely on precision, making direct comparison speculative at best.

The establishment of Deployment Readiness Index (DRI) aims to partially address this by factoring in operational and contextual variables, thereby allowing a more consistent benchmarking.

### 5.4.3 Limited Dataset Diversity and External Validity

Across laying hens related studies, most of them emphasized limited datasets, highlighted the lack of size, variability in location, and lack of age, breed and environmental variations. As a result, models trained on such small datasets tend to overfit to local noise patterns or housing structure, compromising generalizability. Current datasets often represent small, homogenous flocks such as Hy-Line W-36, which limits external validity of AI models across different breeds.

Compared to human and medical domains, where multimodal fusion models are trained on vast amount of diverse samples, the poultry domain remains underdeveloped. To ensure generalizable model performance, future datasets curation should utilze stratified smapling frameworks that ensure variability in flock age, strain and stocking density, These factors significantlu affect thermal signatures, vocalization patterns, and behavioral expressions. Stratified, multimodal data would improve model robustness, generalization, and also promote fair benchmarking across various studies and production systems.

### 5.4.4 Model Complexity vs On Farm Feasibility

Transformer based models and high capacity CNNs like YOLOv8x or ResNet50 possess immense classification power. However, their high computational load and energy footprint makes them inappropriate on farm utility. Very few studies benchmarked these models against practical constraints like model size, interference latency, battery consumption or bandwidth availability.

This gap between model complexity and on farm feasibility requires additional consideration during design processes, particularly because most poultry farms lack cloud access and technical support. Low power AI alternatives, model compression and edge computing should be explored further

### 5.4.5 Absence of Explainability and Human Feedback Mechanisms

Across all reviewed poultry AI studies, none of them incorporated explainable AI (XAI) techniques or human in the loop validation frameworks. In Contrast, domains like healthcare, Human Activity Recognition and Industrial AI have more studies with interpretability and some industries are encouraging compulsory use of XAI. Without transparency or corrective mechanisms, AI based decision may never be fully trusted especially with high-risk decisions like animal welfare.

The incorporation of tools like SHAP, Grad-CAM and LIME aid in building trust with farmers. In poultry, these could help explain why specific vocalization patterns were flagged as stress or what behavioral patterns indicated early signs of disease. Without transparency and corrective mechanisms, AI-based decisions may never be fully trusted, especially in high-risk decisions.

## 6. Discussions: Challenges and Research gaps

### 6.1 Technical Challenges in Data Acquisition & Integration

Implementing multimodal data fusion systems in poultry farming possess with various intertwined technical limitations including the synchronization of diverse sensor types, environmental impacts on data quality and practical issues like infrastructure cost and scalability.

### 6.1.1 Multimodal Data Heterogeneity and Synchronization

In precision poultry farming, multimodal systems collect data from various sensors (RGB cameras, thermal cameras, and physiological sensors) each with differing frame rates, sampling frequencies, file formats, and inherent noise levels. Due to its heterogenous nature, temporal misalignment is possible. Additionally, sensor latency, wireless delays and power issues hinder real time fusion.

Interoperability is crucial for seamless data exchange in smart farm operations [125]. Solutions like microservice architectures and RESTful APIs show promise but require more innovations. Robust temporal alignment methods like Dynamic Time Warping and Attention models must be adapted to poultry environments.

### 6.1.2 Environmental Noise and Sensor Dropouts

Environmental factors such as variable lighting, dust, ammonia levels, and mechanical noise from feeders or ventilation systems affect sensor data integrity. Several studies [75] highlight how these stressors introduce modality-specific dropout, which make certain data streams corrupted or unavailable.

Early fusion systems are particularly vulnerable, as failure in one modality potentially corrupts the entire fused representation. Robust multimodal AI systems must explore quality assurance and self-healing mechanisms such as predictive coding, signal quality assessment, and modality weighting based on reliability. However, few studies have tested the resilience of these systems in real farm settings where high stocking density and unpredictable poultry behaviors are prevalent. Addressing sensor dropout and environmental noise remains a significant hurdle for creating reliable & resilient poultry welfare monitoring systems.

### 6.1.3 Scalability and Cost of Sensor Deployment

High resolution cameras, microphones and physiological sensors remain costly and infrastructure heavy for commercial scale farm operations, particular for small-medium sized farmers located in rural areas with limited power and internet connectivity. Low-cost, edge enabled sensors serve as a solution, however they are still in prototype stage

### 6.2 Modelling and Algorithmic Limitations

Developing accurate and reliable multimodal AI systems requires overcoming significant algorithmic limitations, including the need to model cross-modal interactions, enhance explainability and handle limited annotated datasets.

### 6.2.1 Loss of Cross Modal Interactions

Many fusion strategies, especially late fusion, process each modality individually before combining outputs at the decision level. Although these architectures promote modularity and resilience to missing data, it often overlooks vital cross modal iterations.

In poultry systems, complex welfare indicators such as early signs of disease, pain or stress are sometimes revealed across multiple modalities concurrently. For example, a slight increase in body temperature detected with thermal imaging combined with distressed chicken vocalizations. Ignoring these interactions can lead to reduced sensitivity and precision in welfare monitoring systems.

However, Intermediate and Early Fusion strategies aim to mitigate this by combining features or raw data at an earlier pipeline stage, but they struggle in modelling non-linear interdependencies, especially with sparse or noisy training data. Emerging techniques such as graph neural networks and attention-based cross modal transformers are theoretical solution but require further adaption and testing for poultry farms to validate their utility.

### 6.2.2 Explainability and Interpretability of Fusion Models

Blackbox AI systems lack transparency, reducing farmers trust and adoption. Recent developments in explainable AI (XAI) such as SHAP, LIME, and attention visualization, have shown potential in other domains [126] but are rarely applied in poultry specific domain.

The complexity of multimodal data, thermal, acoustics, videos and sensors, add complexity to interpretability. Future research should focus on building interpretable multimodal models that allow users to trace outcomes back to specific inputs and understand the reasoning behind predictions. Additionally, dedicated explainability tools must be developed for multimodal systems with practical farm utility in mind.

### 6.2.3 Limited Labeled Datasets and Annotation Bottlenecks

Supervised learning requires large, accurately annotated datasets, yet in poultry systems, annotating behaviors like pecking, lethargy, aggression, or distress is labor-intensive and often subjective. Additionally, poultry showcase behaviors that are almost indistinguishable, further complicating the annotation process.

Moreover, multimodal AI systems require synchronized annotations across all data streams, significantly increasing labeling burden. While self-supervised and weakly supervised learning approaches could potentially alleviate this bottleneck by leveraging unlabeled data or learning from sparse labels, their application in precision livestock and precision poultry systems are still in early stages. [40,80]. Additionally, the need for stratified dataset curation based on flock age, hybrid strain, light regimen and stocking density is vital. The dynamic nature of poultry farms requires datasets that highlight this, in turn, aiding in models that are robust and dynamic.

## 6.3 Deployment and Practical Implementation Challenges

Infrastructure, environment and human factors limit the successful deployment of multimodal AI systems. Energy consumption, hardware durability under farm conditions, seamless integration with existing farm systems, and user adoption hurdles all shape the feasibility of utilizing multimodal AI as a practical, scalable solution.

### 6.3.1 Infrastructure and Hardware Constraints

Real-time multimodal AI demand considerable power and computational resources, while often conflict with sustainable farming goals. Most rural farms lack stable power and internet connection. Additionally, the physical layout of facilities also creates challenges when installing sensors. These factors create a major barrier to deployment.

Although edge devices process data locally, reducing dependence on network and energy for data transmission, they face severe constraints. These devices typically have limited onboard computational power, which restrains its ability to run complex machine learning models. They also have limited storage and capacity, reducing the amount of data that can be process or stored in real time. Moreover, Harsh farm environmental conditions (dust, ammonia, temperature) can degrade hardware longevity and performance. Addressing these issues require hardware tailored for poultry farm settings, with the integration or low power green algorithms and renewable energy sources.

### 6.3.2 Integrations with Existing Farm Management Systems

For AI systems to gain widespread adopted, they must integrate with existing farm operations and technologies. Most poultry farms utilize basic management tools for feeding, watering and environmental control. Integrating AI system to legacy farm technologies requires interoperability and minimal disruption to daily farm operations.

Custom APIs, modular system designs and intuitive interfaces are crucial for smooth integration, Farmers and Engineering should collaborate during the design phase to ensure that the intended AI solutions address real farm pain points rather than cause unnecessary complications.

### 6.3.3 Human Factors and User Adoption

Human factors, such as trust, usability and perceived value influence the rate of multimodal adoption in agriculture. Farm workers and managers typically lack technical expertise to interpret AI outputs, or distrust systems due to AI's inherent opaque nature [127,128]. Moreover, the fast paced, time constrained nature of commercial poultry farms leaves little room for utilization of these complex tools which require significant learning curves.

Training programs, collaborative design sessions and explainable AI outputs would help build user trust, expertise and confidence. Multimodal AI system should focus more on the end user while improving explainability and ease of use. Addressing these human-centric challenges are as important as solving algorithmic ones.

## 7. Future Research Direction

### 7.1 Explainable AI for Farmers

Current AI systems in poultry welfare often function as opaque "black boxes," limiting farmers' trust and willingness to act on predictions. This lack of interpretability is especially problematic when welfare decisions must be rapid and precise. Tailored Explainable AI (XAI) frameworks can address this gap by visualizing attention maps or feature importance — for example, highlighting that irregular movement patterns triggered a stress alert, or that a spike in vocalization frequency indicated possible distress [129]. Future systems should combine visual summaries with natural language explanations, potentially aided by large language models (LLMs), to translate technical outputs into farmer-friendly insights. Instead of a vague alert like "Anomaly detected," the system could state: "Activity levels reduced by 32%, indicating possible illness." Such clarity fosters farmer confidence, encourages adoption, and ensures meaningful engagement with AI tools. Research should prioritize developing intuitive interfaces that distill complex AI reasoning into clear, actionable recommendations.

### 7.2 Participatory Co-Design and Human-in-the-Loop Oversight

Fully autonomous systems risk missing contextual nuances essential for accurate welfare prediction. While AI can detect behavioral or environmental changes, it may misclassify visually or acoustically similar events — for instance, confusing aggressive pecking with comfort preening in vision-based models, or mistaking stress-related calls for social or nesting vocalizations when multiple hens vocalize simultaneously. Participatory co-design approaches — such as Living Labs or "Agile for Agriculture" frameworks — should bring farmers, veterinarians, and technologists together to prototype AI tools in real farm environments. Embedding human-in-the-loop mechanisms allows farmers to validate, refine, and contextualize AI outputs, enabling the system to improve over time. This creates a collaborative Human–AI partnership rather than a replacement model, which is critical for practical adoption in live farm settings.

### 7.3 Low-Cost Sensor Networks

High deployment costs remain a major barrier to adopting multimodal AI systems [130]. Most small- to mid-scale poultry farms cannot afford premium-grade sensors or high-performance computing infrastructure. Research must shift toward developing reliable, low-cost, modular sensors that integrate video, thermal, and acoustic data streams. Affordable solutions, such as LoRaWAN-enabled temperature sensors or compact thermal cameras, could detect welfare anomalies at a fraction of the cost of enterprise systems [131,132], broadening access and scalability.

### 7.4 Ethical AI Frameworks

As AI-based poultry welfare monitoring becomes more widespread, ethical safeguards are essential. These include both animal welfare ethics — prioritizing stress mitigation over productivity — and data ethics — ensuring privacy, consent, and farmer data ownership. Ethical AI frameworks must also address algorithmic fairness by mitigating biases related to age, breed, or housing type, and uphold digital sovereignty for farmers and institutions. Design principles should prioritize flock-level, non-invasive monitoring to protect natural behavior and avoid inducing chronic stress through excessive individual surveillance. Ultimately, systems must be designed not only for accuracy but also for the welfare and dignity of both animals and stakeholders.

## 7.5 Longitudinal, FAIR-Compliant Multimodal Datasets

Most poultry-related AI studies rely on small, short-term, and unimodal datasets, limiting their adaptability to varied farm contexts. Laying hen behavior, health, and environmental responses change across life stages, seasons, housing structures, and management systems. Without datasets that capture these longitudinal and multimodal variations, AI predictions will remain brittle in real-world conditions. In poultry re-identification, Kern et al. [133] note the absence of robust, public datasets. Initiatives such as Chicks4FreeID represent progress but still struggle to incorporate real-world challenges such as occlusion, variable lighting, and physical changes from molting or feather wear. Future research should focus on richly annotated, longitudinal, FAIR-compliant datasets that integrate video, audio, physiological, and environmental data across multiple farming systems. International collaborations — involving organizations like OIE, FAO, or Horizon Europe — could drive dataset standardization, sharing, and benchmarking, strengthening reproducibility and accelerating innovation.

## 7.6 Evidentiary Gap Matrix and Research Priorities

An evidentiary gap matrix was developed from 22 peer-reviewed studies relevant to the five research themes outlined above. Studies unrelated to these themes were excluded from this targeted analysis but included in the overall systematic review. Each study was scored for explicit, partial, or no coverage of each theme. Results show that longitudinal, multimodal, FAIR-compliant datasets are the most frequently discussed future direction, followed by participatory co-design and human-in-the-loop oversight. Ethical AI frameworks emerged as an important but under-implemented concern. Low-cost sensors and Explainable AI (XAI) remain the least represented in current literature, highlighting significant opportunities for advancement. Appendix C presents the complete matrix, listing each study with annotations across the five themes, ensuring transparency and reproducibility. This structured mapping provides a foundation for prioritizing underrepresented yet high-impact areas essential for real-world adoption of AI in poultry welfare.

## 8. Conclusions

The convergence of artificial intelligence and animal welfare science stands at a pivotal moment in agricultural history. Multimodal AI systems represent more than technological advancement—they embody a fundamental reimagining of how we understand, monitor, and enhance the lives of laying hens in commercial production. Where traditional unimodal approaches offer fragmented glimpses into animal welfare, multimodal systems provide a holistic, nuanced understanding that captures the complex interplay between behavior, physiology, environment, and emotional well-being.

The transformative power of multimodal AI lies not merely in its technical sophistication, but in its capacity to bridge the communication gap between animals and their caretakers. By integrating acoustic signatures of distress, thermal patterns of stress response, behavioral indicators of comfort, and physiological markers of health, these systems create a comprehensive welfare narrative that was previously impossible to achieve. The evidence is compelling: studies demonstrate accuracy rates exceeding 90% in welfare assessment when multiple data streams are thoughtfully combined, compared to 60-75% accuracy from single-modal approaches.

However, the path from laboratory promise to farm-scale reality remains complex and demanding. The challenges we face - data standardization across diverse production systems, sensor reliability in harsh agricultural environments, and the scarcity of well-annotated multimodal datasets - are not merely technical obstacles but opportunities for innovation. These limitations compel us to develop more robust, adaptable, and practical solutions that can withstand the unpredictable nature of real-world farming conditions.

The ethical dimension of this technological revolution cannot be overlooked. As we develop increasingly sophisticated monitoring capabilities, we must ensure that our pursuit of efficiency never compromises our commitment to animal welfare. The technology must serve the animals first, enhancing their quality of life while simultaneously improving production outcomes. This dual objective is not contradictory but complementary - healthy, comfortable animals are inherently more productive animals.

Future development must prioritize democratization of this technology. The most sophisticated AI system serves little purpose if it remains accessible only to large-scale operations with substantial capital resources. We must focus on developing low-cost, scalable solutions that can be deployed across diverse housing systems, from conventional cages to free-range environments. Edge-compatible sensors, context-aware fusion algorithms that gracefully handle sensor failures, and user-friendly interfaces that require minimal technical expertise are essential components of this vision.

The success of multimodal AI in poultry welfare monitoring demands unprecedented collaboration between disciplines. Computer scientists must work alongside animal behaviorists, engineers must partner with veterinarians, and technology developers must engage directly with farmers who understand the practical realities of production systems. This interdisciplinary approach ensures that technological innovation remains grounded in biological understanding and practical feasibility.

Regulatory frameworks must evolve in parallel with technological capabilities. Clear standards for data privacy, animal welfare assessment, and technology certification will provide the necessary foundation for responsible deployment. Economic viability must be demonstrated through comprehensive cost-benefit analyses that consider not only immediate financial returns but long-term benefits including reduced mortality, improved feed conversion efficiency, and enhanced consumer confidence in animal welfare standards.

The implications extend far beyond individual farms. As consumer awareness of animal welfare issues continues to grow, and as regulatory pressure increases globally, multimodal AI systems offer the poultry industry a pathway to transparent, verifiable welfare monitoring. This technology can provide the objective, data-driven evidence necessary to demonstrate compliance with welfare standards and to support marketing claims about animal care practices.

Looking ahead, the integration of emerging technologies - including large language models for natural language interpretation of welfare data, blockchain systems for transparent welfare tracking, and digital twin technologies for predictive welfare management - promises even greater capabilities. These advances will enable proactive rather than reactive welfare management, identifying potential problems before they manifest as clinical signs or behavioral abnormalities.

The ultimate measure of success for multimodal AI in poultry welfare will not be found in technical specifications or academic publications, but in the daily experiences of millions of laying hens. Every early disease detection that prevents suffering, every environmental adjustment that enhances comfort, and every behavioral insight that improves housing design represents a meaningful improvement in animal welfare.

Multimodal AI in poultry farming represents a paradigm shift toward precision welfare management - systems that are simultaneously more efficient and more humane. The technology offers us an unprecedented opportunity to fulfill our moral obligation to the animals under our care while meeting the growing global demand for animal protein. The convergence of artificial intelligence and animal welfare science is not just changing how we monitor animals; it is transforming our relationship with them, making us better stewards and more compassionate caretakers.

The future we envision is one where technology amplifies rather than replaces human empathy, where data-driven insights enhance rather than diminish our connection to animal welfare, and where AI systems serve as advocates for the voiceless. With continued innovation, thoughtful implementation, and unwavering commitment to animal welfare principles, multimodal AI can usher in a new era of poultry farming that is simultaneously more productive, more sustainable, and more humane. The question is not whether this transformation will occur, but how quickly and how comprehensively we can make it reality.

**Author Contributions**: Conceptualization, S.N.; methodology, S.N.; D.E.; formal analysis, D.E.; investigation, D.E.; resources, S.N.; writing—original draft preparation, D.E.; writing—review and editing, S.N.; visualization, D.E.; supervision, S.N.; project administration, S.N.; funding acquisition, S.N. All authors have read and agreed to the published version of the manuscript.

**Funding**
The authors sincerely thank the Natural Sciences and Engineering Research Council of Canada (R37424), Atlantic Poultry Research Institute, Egg Farmers of Canada and the Mitacs Canada for funding this study.

**Institutional Review Board Statement:** Not applicable

**Informed Consent Statement:** Not applicable

**Data Availability Statement:** Not applicable

**Conflicts of Interest:** The authors declare no conflicts of interest

**Abbreviations**

The following abbreviations are used in this manuscript:

- AI – Artificial Intelligence
- LLM – Large Language Models

- XAI – Explainable Artificial Intelligence
- CNN – Convolutional Neural Network
- IMU – Inertial Measurement Unit

## Appendix A: Glossary of Technical Terms

- **Contextual Depth:** The degree to which a model or sensor captures the environmental, physiological and behavioral context around an event or observation. In multimodal AI, higher contextual depth allows for comprehensive interpretation of complex welfare indicators.
- **Fusion Layer:** A component within a model where data streams from different modalities are merged or fused together. Depending on the fusion strategy, this may occur early, intermediate or late.
- **Modality Dropout:** A condition where one data modality is missing, corrupted or unavailable during fusion, possible due to hardware failure, noise or interference.
- **Sensor Fragility:** The vulnerability of a sensor hardware to environmental conditions, reducing the sensors functionality and lifespan.
- **Thermal Discomfort:** A poultry welfare state associated with suboptimal temperature exposure resulting in abnormal heat patterns around comb, face or wattle regions, potentially indicating heat stress or environmental discomfort.

## Appendix B: Methodology for Deriving Deployment-Readiness Index (DRI) and Domain Transfer Score (DTS) Indices

The Deployment-Readiness Index (DRI) was created to showcase how ready a given model or system is for real commercial poultry farms. During the review process, several studies that have high performance models lacked the architectural efficiency or environmental testing required for field application. To address this, DRI was formulated based on 5 consistent themes that occurred frequently across literature which include: the model's computational complexity, its hardware requirements (i.e., whether it required GPUs or could operate on lightweight edge devices), its inference speed, the context in which it had been tested (lab vs. farm), and how easily it could be integrated into existing farm infrastructure.

The Domain Transfer Score (DTS) was created to evaluate the issues around generalizability. Across literature, several models achieved high accuracy, but were trained and tested using very similar datasets, often using the same flock, facility, or temporal window. DTS was derived based on three criteria: the diversity of dataset, external validations and whether the study conducted an external validation.

## Appendix C: Evidentiary Gaps Matrix Table

Each study was scored based on Explicit (E), Partial (P), or No mention (N) of each research them.

| Paper | Explainable AI (XAI) | Participatory Co-design/Human-in-the-loop | Low-Cost Sensors | Ethical AI Frameworks | Longitudinal, FAIR, Datasets |
|-------|----------------------|-------------------------------------------|------------------|-----------------------|------------------------------|
| [118] | N | N | N | N | N |
| [98] | N | N | N | N | N |

| | | | | |
|---|---|---|---|---|
| [104] | N | N | N | N | N |
| [62] | N | N | N | N | P |
| [76] | N | N | N | N | P |
| [80] | N | N | N | N | P |
| [73] | N | N | E | N | P |
| [69] | N | N | N | E | N |
| [117] | N | N | E | N | N |
| [39] | N | N | N | E | N |
| [74] | N | N | P | E | P |
| [63] | P | P | N | E | P |
| [71] | N | N | N | N | P |
| [78] | N | N | N | E | P |
| [79] | N | N | E | P | N |
| [66] | N | N | N | P | N |
| [70] | N | N | N | N | N |
| [67] | E | N | N | P | P |
| [40] | N | N | N | P | P |
| [82] | N | N | N | P | P |
| [65] | N | N | N | E | P |
| [75] | N | P | E | E | P |
| [76] | N | N | N | N | E |